%% file: main.tex
\documentclass[10pt,twocolumn,letterpaper]{article}

\usepackage{cvpr}              


\definecolor{cvprblue}{rgb}{0.21,0.49,0.74}
\usepackage[pagebackref,breaklinks,colorlinks,allcolors=cvprblue]{hyperref}

\usepackage{newtxtext}
\usepackage{amsmath}
\usepackage{amssymb}
\usepackage{amsfonts}
\usepackage{multirow}
\usepackage{booktabs}
\usepackage{tabularx}
\usepackage{enumitem}
\usepackage{epstopdf}
\usepackage[labelfont=bf,textfont=bf]{caption}

\usepackage{algorithm}  
\usepackage{bbm}
\usepackage{makecell}
\usepackage{bm}
\usepackage{pifont}
\usepackage{bbding}
\usepackage{color}
\usepackage{colortbl}

\usepackage{tikz}                           
\usepackage{pgfplots}                       
\pgfplotsset{compat=1.17}                   

\definecolor{jmyellow}{RGB}{255,215,0}      
\definecolor{jmlight}{RGB}{255,245,200}     

\newcommand{\stitle}[1]{\vspace{0.8mm} \noindent {\bf #1}}

\definecolor{mycustompurple}{RGB}{154, 36, 79} 
\usepackage[utf8]{inputenc}
\usepackage{hyperref} 
\hypersetup{
    colorlinks=true,            
    linkcolor=red,              
    filecolor=mycustompurple,   
    urlcolor=magenta            
}

\def\paperID{6674} 
\def\confName{CVPR}
\def\confYear{2026}

\title{GrOCE~\hspace{-0.2em}\raisebox{-0.3ex}{\includegraphics[width=1.2em,height=1.2em]{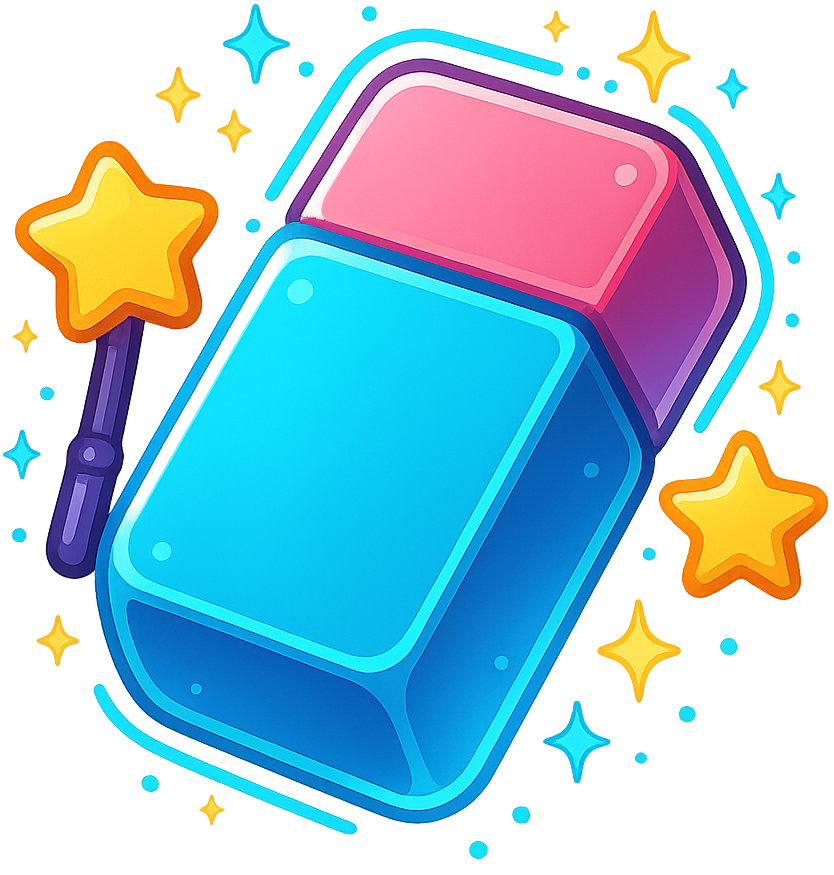}}: Graph-Guided Online Concept Erasure for Text-to-Image Diffusion Models}

\author{
Ning Han\textsuperscript{1} \quad
Zhenyu Ge\textsuperscript{1} \quad
Feng Han\textsuperscript{2} \quad
Yuhua Sun\textsuperscript{1} \quad
Chengqing Li\textsuperscript{1} \quad
Jingjing Chen\textsuperscript{2}\thanks{Corresponding author.} \\[0.5em]  
\textsuperscript{1}School of Computer Science, Xiangtan University \\[0.3em]
\textsuperscript{2}School of Computer Science, Fudan University \\[0.5em]
{\tt\small \{hanninginf,gezhenyu12,drnatsun,DrChengqingLi\}@gmail.com,} \\
{\tt\small fhan25@m.fudan.edu.cn, chenjingjing@fudan.edu.cn}
}

\begin{document}
\maketitle

\begin{abstract}

Concept erasure aims to remove harmful, inappropriate, or copyrighted content from text-to-image diffusion models while preserving non-target semantics. However, existing methods either rely on costly fine-tuning or apply coarse semantic separation, often degrading unrelated concepts and lacking adaptability to evolving concept sets. In this paper, we propose \textbf{Gr}aph-Guided \textbf{O}nline \textbf{C}oncept \textbf{E}rasure (GrOCE), a training-free framework that performs precise and context-aware online removal of target concepts. GrOCE constructs dynamic semantic graphs to identify clusters of target concepts and selectively suppress their influence within text prompts. It consists of three synergistic components: (1) dynamic semantic graph construction  (\textsc{Construct}) incrementally builds a weighted graph over vocabulary concepts to capture semantic affinities; (2) adaptive cluster identification (\textsc{Identify}) extracts a target concept cluster through multi-hop traversal and diffusion-based scoring to quantify semantic influence; and (3) selective severing (\textsc{Sever}) removes semantic components associated with the target cluster from the text prompt while retaining non-target semantics and the global sentence structure. Extensive experiments demonstrate that GrOCE achieves state-of-the-art performance on the Concept Similarity (CS) and Fréchet Inception Distance (FID) metrics, offering efficient, accurate, and stable concept erasure. {\underline{Our code is available at this \href{https://github.com/MagicCat-AI/GrOCE}{link}}.}
\end{abstract}


\begin{figure*}
    \centerline{\includegraphics[width=0.97\textwidth]{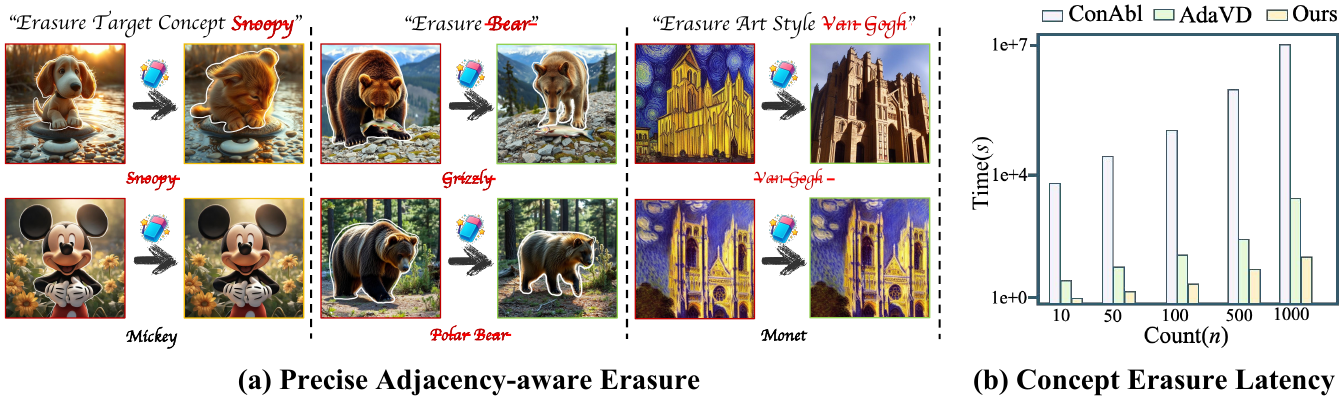}}
\caption{
Two key aspects of concept erasure.
(a) Concept erasure in text-to-image diffusion models involves both explicit and implicit semantic structure in the latent space. Our method leverages adjacency in semantic space to suppress a target concept while better preserving its neighboring, non-target concepts.
(b) Runtime comparison with the training-based ConAbl~\cite{kumari2023ablating} and the recent training-free AdaVD~\cite{wang2025precise}. Our method achieves an order-of-magnitude speedup, making online large-scale concept removal practical.}
\label{fig:motivation}
\end{figure*}

\section{Introduction}\label{sec:intro}

Text-to-image diffusion models~\cite{ho2020denoising,dhariwal2021diffusion,nichol2021improved,rombach2022high,zhang2023adding} have rapidly evolved into the dominant paradigm for controllable image generation. Despite their remarkable capabilities, diffusion models frequently produce harmful, biased, or copyright-infringing content, raising serious ethical and regulatory concerns~\cite{somepalli2023diffusion,leu2024auditing,yang2024sneakyprompt,zeng2024advi2i,zhu2025reinforcing}. In response, existing concept erasure methods typically rely on fine-tuning model parameters and designing carefully crafted erasure objectives to achieve the desired effect~\cite{lu2024mace,gong2024reliable,lyu2024one,wang2025precise,lee2025localized}. However, they are unable to remove newly emerging concepts online. As diffusion models increasingly serve as a cornerstone for creative industries, digital media, and design automation, undesirable or copyright-protected concepts can emerge unpredictably, making it impossible to maintain a complete predefined list of concepts to remove. Consequently, concept erasure has emerged as a critical research area for ensuring the safety and reliability of generative diffusion models. Achieving online removal of harmful semantic concepts while preserving non-target content remains a significant technical challenge.

Diffusion models present fundamental challenges for concept erasure. Unlike discriminative models with discrete decision boundaries, diffusion models distribute concept representations across temporal denoising trajectories, spatial attention maps, and cross-modal embeddings simultaneously. This results in deeply entangled representations: modifying one concept (e.g., violence) can unintentionally alter semantically adjacent or visually similar concepts (e.g., conflict, action, or even intensity). The iterative nature of diffusion models compounds this challenge: perturbations introduced early in the denoising process amplify exponentially, leading to either incomplete erasure or widespread semantic damage. This characteristic renders simple keyword-based filtering methods ineffective for concept removal in diffusion models. 

Existing concept erasure methods fall into two main categories, each with critical drawbacks. Early solutions finetune or modify the weights of pre-trained models to erase target concepts  \cite{gandikota2023erasing,heng2023selective,gandikota2024unified,kumari2023ablating,fan2023salun}. 
For example, ESD \cite{gandikota2023erasing} and CA \cite{kumari2023ablating}, align the probability distributions of the targeted concept with that of a
null string. Nonetheless, this incurs high computational cost, suffers from catastrophic forgetting, and struggles to adapt to emerging risks. Later work (e.g., MACE \cite{lu2024mace}, SPM \cite{lyu2024one} and CPE \cite{lee2025concept}) mitigates the problem of catastrophic forgetting by introducing lightweight adapters and regularization mechanisms to decouple unsafe concepts, yet still requires substantial fine-tuning time. To overcome this limitation, inference-time interventions such as SPEED~\cite{li2025speed}, UCE~\cite{gandikota2024unified}, and AdaVD \cite{wang2025precise} apply on-the-fly edits to activations or attention maps. These methods are efficient but rely on heuristic mappings that fail to capture deeper semantic entanglements, making accurate boundary detection elusive. As illustrated in Figure \ref{fig:motivation}, removing “Bear” erases its explicit features yet leaves related notions like “grizzly” and “polar bear” intact. Crucially, all prior approaches treat concepts in isolation, they cannot adapt online to evolving concept sets and ignore the rich relational structure of the latent semantic space.

In this paper, we propose \textbf{Gr}aph-Guided \textbf{O}nline \textbf{C}oncept \textbf{E}rasure (GrOCE), a training-free framework that performs precise, context-aware online removal of target concepts by leveraging dynamic semantic graphs to identify concept clusters and selectively suppress their influence in the prompt embeddings. As shown in Figure~\ref{fig:framework}, the framework consists of three synergistic components: (1) Dynamic Semantic Graph Construction (\textsc{Construct}), which builds real-time semantic graphs over prompt embeddings; (2) Adaptive Cluster Identification (\textsc{Identify}), which identifies target concept clusters via multi-hop traversal and diffusion-based scoring; and (3) Selective Severing (\textsc{Sever}), which selectively removes the influence of the identified concept cluster while preserving non-target semantics and the global structure of the prompt. GrOCE provides three key advantages (see Figure~\ref{fig:motivation}). First, it operates entirely at inference time, requiring no gradient access or retraining, which enables real-time adaptation to emerging erasure needs. Second, its graph-based formulation captures multi-concept entanglement and higher-order dependencies that token-level heuristics fail to model. Third, the explicit semantic graph enhances interpretability by revealing not only what is erased but also why. 

Our overall contributions are thus three-fold: i) we propose GrOCE, a graph-guided method for online concept erasure in text-to-image diffusion models, enabling structured reasoning over semantic dependencies; ii) we develop a training-free framework that leverages dynamic semantic graphs to detect and selectively suppress target concepts in a context-aware manner; and iii) we conduct extensive experiments across multiple tasks, including cartoon concept removal and artistic style erasure, demonstrating that GrOCE achieves state-of-the-art performance in erasure accuracy, non-target fidelity, and inference efficiency.

\section{Related Work}\label{sec:related} 
Recent advances in concept erasure aim to enhance the safety of diffusion models by removing targeted concepts while preserving generative utility. However, most existing methods adopt a \emph{local} view, treating concepts as isolated entities that can be surgically removed~\cite{schramowski2023safe,orgad2023editing,belrose2023leace,han2025dumo,biswas2025cure}. This oversimplifies how semantics are encoded and helps explain recurrent failures in practice, motivating a more topological perspective. We categorize existing concept-erasure methods into two paradigms, each revealing deeper limitations of this local view.

\noindent
\textbf{Parameter-Based Concept Erasure.} Early studies suggest that concepts can be surgically removed from model weights~\cite{gandikota2023erasing,huang2024receler,bui2024erasing,thakral2025fine,li2025sculpting}. Erased Stable Diffusion (ESD)~\cite{gandikota2023erasing} pioneered fine-tuning with adversarial objectives, while EraseAnything~\cite{gao2025eraseanything} adopted LoRA-based parameter adjustments and attention-map regularizers to selectively suppress unwanted activations. SPM~\cite{lyu2024one} and FMN~\cite{yang2022focal} refined loss functions for better precision. However, these methods implicitly adopt an oversimplified view of concept encoding: diffusion models represent concepts as \emph{entangled manifolds with fuzzy boundaries}~\cite{stanczuk2024diffusion}. Attempts to erase ``Van Gogh'' style tend to damage ``impressionism'' and ``brushstroke'' due to overlapping parameter subspaces, making perfectly surgical removal infeasible in practice.

\noindent
\textbf{Inference-Time Interventions.} Inference-time interventions avoid retraining by modifying attention patterns or a small set of parameters at test time to suppress specific concepts \cite{schramowski2023safe,zhang2024generate,gandikota2024unified,li2025speed,li2025detect}. Representative approaches include UCE~\cite{gandikota2024unified}, SLD \cite{schramowski2023safe}, RECE~\cite{gong2024reliable}, SPEED~\cite{li2025speed}, and Forget-Me-Not~\cite{zhang2024forget}, which apply closed-form or optimization-based updates to cross-attention or lightweight parameter subsets to steer generation away from unwanted concepts while preserving overall capability. A common limitation of these methods is their assumption that target concepts manifest as stable, locally suppressible activation patterns. In practice, activations are \emph{highly sensitive to context and sampling dynamics}: paraphrased or compositional prompts, as well as minor changes to the denoising schedule, can route information through alternative pathways, causing erasure effectiveness to vary across prompts and data distributions. For example, a UCE~\cite{gandikota2024unified} configuration that suppresses ``violence'' for one prompt template may fail under simple rephrasings. These instabilities limit the reliability of such interventions in safety-critical settings. Additionally, several recent methods perform concept erasure without retraining by manipulating text embeddings or attention maps, including AdvUnlearn~\cite{zhang2024defensive}, Getwant~\cite{li2024get}, TRCE~\cite{chen2025trce}, and AdaVD~\cite{wang2025precise}. However, they remain constrained by local assumptions about concept geometry, limiting their ability to handle overlapping or complex concepts. For example, TRCE~\cite{chen2025trce} cannot address visual-only concepts (e.g., pose-lighting interactions), and AdaVD~\cite{wang2025precise} assumes linear separability, which fails for non-convex concept regions (e.g., “weapon” vs. “tool”). Most critically, these methods overlook the topological structure of concept relationships, such as similarity neighborhoods and co-occurrence patterns, which our graph-guided approach seeks to address.

Building on these observations, we adopt a \emph{graph-guided} perspective, modeling concepts as an interconnected graph of similarity, hierarchy, and co-occurrence. Rather than treating concepts in isolation, our framework explicitly captures their interdependencies. We instantiate this perspective through a \emph{training-free, online} procedure that retains the practical advantages of inference-time methods while improving robustness to prompt/timestep variations and reducing collateral damage to semantically adjacent concepts.

\section{Preliminary}

Let $\mathcal{D}: \mathcal{T} \rightarrow \mathcal{I}$ be a pre-trained text-to-image diffusion model that maps a text prompt $t \in \mathcal{T}$ to an image in the output space $\mathcal{I}$, where the prompt $t$ consists of $M$ words, i.e., $t = \{t_1, \dots, t_M\}$. Given a target concept $c_t$ to erase, our goal is to construct a modified model $\mathcal{D}'$, defined as 
\begin{equation}
\mathcal{D}'(t) = 
    \begin{cases}
    \mathcal{D}(t)      & \text{if } c_t \notin \text{sem}(t); \\
    \mathcal{D}(\theta(t)) & \text{if } c_t \in \text{sem}(t),
\end{cases}
\end{equation}
where $\mathrm{sem}(t)$ denotes the semantic content embedded in the prompt $t$, and $\theta(\cdot)$ is a concept erasure operator that removes all traces of the target concept $c_t$ while preserving the orthogonal (non-target) semantics.

The key idea of GrOCE is to leverage a dynamic semantic graph $\mathcal{G} = (\mathcal{V}, \mathcal{E}, \mathcal{W}, X)$ for the concept erasure task. Here, $\mathcal{V}$ denotes the set of nodes, $\mathcal{E}$ denotes the set of edges, $\mathcal{W}$ denotes the set of edge weights, and $X \in \mathbb{R}^{|\mathcal{V}| \times D}$ is the concept embedding matrix, where each row corresponds to the embedding of a node in $\mathcal{V}$. The concept embeddings in $X$ are derived from a discrete concept vocabulary, i.e., the set of token-level semantic units defined by the pretrained text encoder of the diffusion model~\cite{radford2021learning}. These embeddings serve as the foundation for constructing semantic relationships among concepts. Within this graph framework, concept erasure can be interpreted as a graph cut problem. The goal is to identify a minimal subset of vertices $\mathcal{V}_{c} \subset \mathcal{V}$ whose removal eliminates all semantic paths to the target concept $c_t$ while preserving connectivity among non-target concepts. In the following, we present our proposed solution to this problem.

\section{Methodology}

\begin{figure*}
    \centerline{\includegraphics[width=0.98\textwidth]{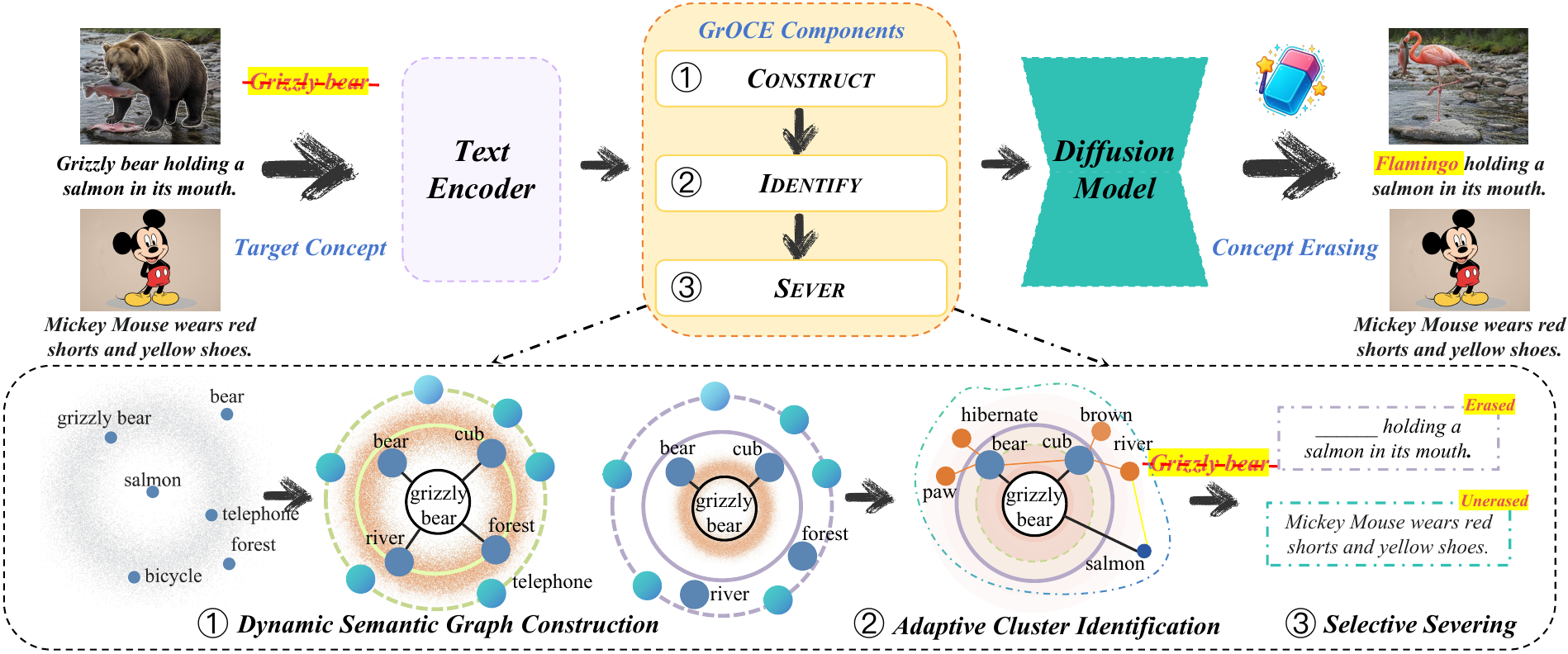}}
\caption{
The GrOCE pipeline for online concept erasure. Given a text prompt and a specified target concept (e.g., “bear”), GrOCE performs inference-time concept erasure through three synergistic components: (1) Dynamic Semantic Graph Construction builds a semantic graph with vocabulary tokens as nodes and cosine-weighted edges, supporting incremental updates for evolving concept sets.(2) Adaptive Cluster Identification performs multi-hop traversal with similarity decay to identify semantically entangled concepts (e.g., “grizzly,” “panda”) around the target. (3) Selective Severing removes the semantic components associated with the identified cluster, editing the text prompt prior to diffusion to suppress target concepts while preserving non-target semantics.
}
\label{fig:framework}
\end{figure*}

The overall pipeline of GrOCE is illustrated in Figure~\ref{fig:framework}. It is a training-free framework that performs concept erasure as an online inference process over dynamic semantic graphs in the representation space. The framework consists of three key components (Secs.~\ref{sec:3.2}--\ref{sec:3.4}): (1) \textsc{Construct} (Sec.~\ref{sec:3.2}), which builds graphs from contextualized embeddings to capture latent concept relationships beyond surface-level similarity; (2) \textsc{Identify} (Sec.~\ref{sec:3.3}), where spectral analysis is used to detect compact clusters that collectively represent the target concept; and (3) \textsc{Sever} (Sec.~\ref{sec:3.4}), which suppresses the target clusters via a graph-guided soft projection while approximately preserving orthogonal semantic directions.

\subsection{Dynamic Semantic Graph Construction (\textsc{Construct})}
\label{sec:3.2}

\stitle{Formulation.} To enable structure-aware concept erasure, GrOCE first constructs a dynamic semantic graph $\mathcal{G} = (\mathcal{V}, \mathcal{E}, \mathcal{W}, X)$. Each node $v_i \in \mathcal{V}$ corresponds to a concept embedding $x_i \in X$. Each edge $e_{ij} \in \mathcal{E}$ represents a semantic association, and the corresponding weight $w_{ij} \in \mathcal{W}$ quantifies the contextual affinity between the two nodes. This graph provides the foundation for reasoning over semantic dependencies and guides targeted concept isolation in subsequent stages.

\stitle{Edge Construction.} We connect two nodes if their cosine similarity exceeds a local threshold $\tau_i$, and assign a soft weight. It is formulated as
\begin{equation}
w_{ij} = \begin{cases}
\exp\left( -\frac{\tau_{i} - \langle x_i, x_j \rangle}{\sigma} \right), & \text{if } \langle x_i, x_j \rangle > \tau_{i}; \\
\quad \quad \quad \quad 0, & \text{otherwise},
\end{cases}
\end{equation}
where $\sigma$ controls the decay rate of the edge weight near the threshold. This procedure converts the continuous embedding space into a discrete graph, making implicit relationships, such as the hierarchical connection between “grizzly bear” and “bear,” explicit and traversable.

Since semantic density varies across the embedding space, we refine edge thresholds using local similarity and local density, formulated by
\begin{equation}
\tau_{i} = \tau_0 + \lambda \cdot \sqrt{ \frac{1}{|\mathcal{N}_i|} \sum_{j \in \mathcal{N}_i} \left( \langle x_i, x_j \rangle - \mu_i \right)^2 },
\end{equation}
where $\tau_0$ is a base threshold, $\lambda>0$ scales the local similarity variance, the local similarity $\mu_i$ and the neighborhood $\mathcal{N}_i$ are defined as
\begin{equation}
\mu_i = \frac{1}{|\mathcal{N}_i|} \sum_{j \in \mathcal{N}_i} \langle x_i, x_j \rangle,
\quad
\mathcal{N}_i = \left\{ j \mid w_{ij} > 0 \right\}.
\end{equation}

Intuitively, dense node regions in the semantic graph adopt higher thresholds to separate closely related concepts, while sparse node regions use lower thresholds to maintain connectivity. As a result, the constructed graph naturally adapts to local concept density and preserves the overall topology across different semantic zones. Besides, the graph may contain a substantial number of concepts unrelated to the target. Our \textsc{Identify} ensures that only concepts strongly associated with the target are selected, while the subsequent selective severing removes the target concepts and preserves all remaining ones.

\subsection{Adaptive Cluster Identification (\textsc{Identify})} \label{sec:3.3}

\textsc{Identify} can be taken as a pre-processing component of \textsc{Sever}. Without first identifying the concept cluster strongly associated with the target through clustering, the subsequent removal of the target concept via \textsc{Sever} may inadvertently eliminate additional unrelated concepts. To this end, we propose \textsc{Identify} for the target concept. The identification process proceeds as follows: 1) identifying the node corresponding to the target concept using anchor initialization, 2) quantifying the influence of each concept via our proposed semantic diffusion, and 3) forming the final target concept cluster through clustering.

\stitle{Anchor Initialization.} We first locate the node corresponding to the target concept $c_t$. If $c_t$ already exists in the node set $\mathcal{V}$, it is used directly; otherwise, a new node for $c_t$ is inserted into the graph $\mathcal{G}$. This node then serves as the anchor for subsequent operations.

\stitle{Semantic Diffusion.} To quantify the influence range of the target concept while reducing noise from distant nodes in the global graph, we restrict semantic diffusion to an $n$-hop neighborhood around the anchor node $c_t$. Specifically, we construct a local subgraph $\mathcal{G}_{\text{sub}}= (\mathcal{V}_{sub}, \mathcal{E}_{sub}, \mathcal{W}_{sub}, X)$ by collecting all nodes whose shortest-path distance to $c_t$ does not exceed $n$. Formally,
\begin{equation}
\mathcal{V}_{sub} =
\{ v_i \in \mathcal{V} \mid d(v_i,c_t) \le n \},
\end{equation}
where $d(\cdot,\cdot)$ denotes the shortest-path distance on the semantic graph $\mathcal{G}$. 

Subsequently, we simulate a diffusion process over this local subgraph $\mathcal{G}_{sub}$ defined as
\begin{equation}
s = \exp(-\varphi \mathcal{L}_{sub})Y, 
\end{equation}
where $\mathcal{L}_{sub} = I - D_{sub}^{-\frac12} W_{sub} D_{sub}^{-\frac12}$ denotes the normalized graph Laplacian of the subgraph $\mathcal{G}_{sub}$. Here $W_{sub}$ is the adjacency matrix of the subgraph and $D_{sub}$ is the corresponding degree matrix $D_{sub}(i,i) = \sum_j W_{sub}(i,j)$. $Y$ denotes the one-hot vector corresponding to the anchor node $c_t$, and $\varphi$ controls the diffusion scale.

Intuitively, nodes that are closer to $c_t$ in the local subgraph receive higher activation, while distant nodes in the subgraph fade out exponentially. This process allows indirect relationships to emerge naturally through multiple paths. We further perform min-max normalization on the diffusion scores $s$ to scale them to the range $[0,1]$ for better distinguishability, without changing their relative order.

\stitle{Clustering.} We construct the target concept cluster by selecting the top-$K$ nodes with the highest diffusion scores in the local subgraph, where $K$ controls the semantic coverage of the cluster. 
Let $\text{rank}(s_i)$ denote the descending rank of node $v_i$ according to $s_i$. The cluster is defined as
\begin{equation}
\mathcal{V}_{c} =
\{c_t\} \cup 
\{v_i \in \mathcal{V}_{sub} \mid \text{rank}(s_i) \le K\}.
\end{equation}

This procedure yields a target concept cluster $\mathcal{V}_{c}$ that captures concepts closely related to the target while preserving the structural connectivity of the semantic graph. The resulting cluster then serves as input to the \textsc{Sever} component.

\subsection{Selective Severing (\textsc{Sever})} \label{sec:3.4}

\begin{figure*}
    \centerline{\includegraphics[width=0.98\textwidth]{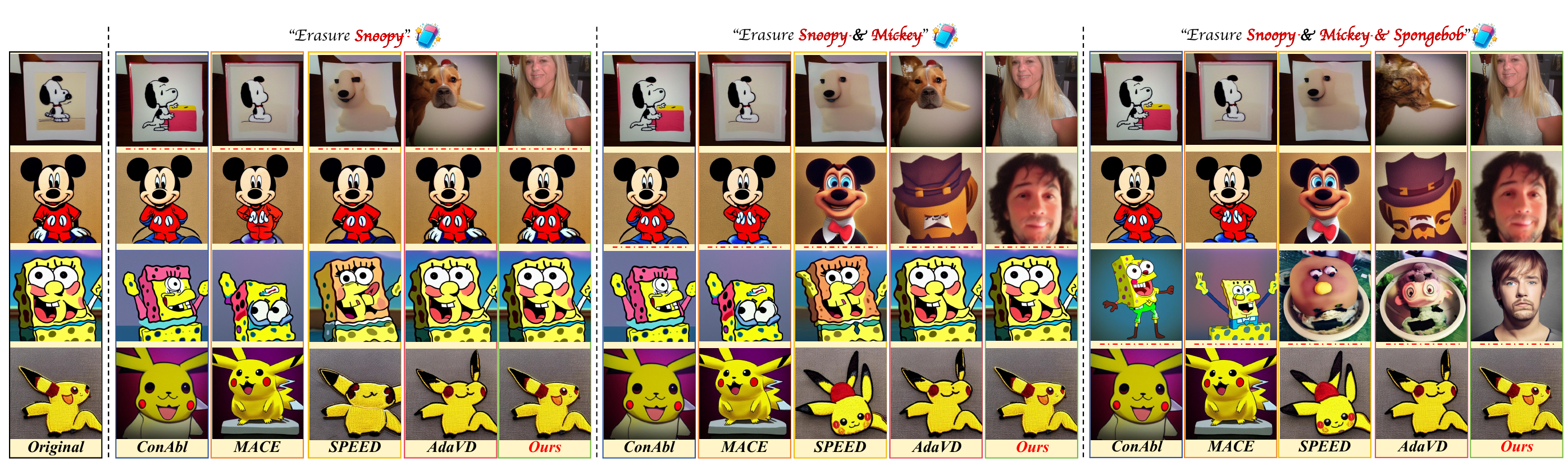}}
\caption{From the visualization results, our method demonstrates excellent erasure and retention capabilities, whether it is erasing Snoopy, Snoopy and Mickey, or Snoopy, Mickey and Spongebob. It can not only accurately accomplish target erasure but also stably retain prior knowledge in the process, thus achieving a balance between effectiveness and information retention.}
\label{fig:concept erasure}
\end{figure*}

Once the target concept cluster $\mathcal{V}_{c}$ has been identified, \textsc{Sever} is applied to eliminate the semantic influence of these concepts from the original text prompt while preserving all remaining semantics. The text prompt $t = [t_1, \dots, t_M]$ is first encoded into token embeddings $f = [f_1, \dots, f_M]$, with $f_i \in \mathbb{R}^D$, using the pretrained text encoder of the diffusion model. For each token embedding $f_i$ corresponding to token $t_i$, we compute its total influence score of token $t_i$  by aggregating contributions from all identified target concepts:
\begin{equation}
\alpha_{i} = \sum_{c'_t\in\mathcal{V}_c}\exp(-\gamma\, d(t_i,c'_t)) \langle f_i, x_j \rangle x_j,
\end{equation}
where $x_j$ is the embedding of concept $c'_t$, $d(\cdot,\cdot)$ denotes the shortest-path distance between token $t_i$ and the concept node $c'_t$ in the semantic graph $\mathcal{G}$, and $\gamma > 0$ controls the decay of influence with distance.

A projection threshold $\delta$ is then applied to determine whether a token should be retained:
\begin{equation}
t' = \left\{ t_i \in t \mid \mathbb{I}\left(\|\alpha_i\| \leq \delta\right) \right\},
\end{equation}
where $\|\cdot\|$ denotes the L2 norm, representing the overall magnitude of the aggregated projection vector for each token. The modified prompt $t'$ is re-encoded to produce refined token embeddings for diffusion generation. The resulting embeddings preserve the global sentence structure and non-target semantics while eliminating the influence of the target concept, enabling precise and context-aware concept erasure guided by the semantic topology of $\mathcal{G}$.

\section{Experiments}

We evaluate GrOCE across various concept erasure tasks, including single-target, multi-target, and art style concept erasure. Our experiments assess both erasure completeness and non-target preservation, with further analysis on scalability, runtime efficiency, hyperparameter sensitivity and robustness discussion.

\subsection{Experimental Setup}

\stitle{Implementation Details.} We implement GrOCE on Stable Diffusion\cite{rombach2022high} using the DPM-solver\cite{lu2022dpm} sampler with 30 sampling steps and classifier-free guidance of 7.5. The hyperparameters are set as follows: the base threshold $\tau_0 = 0.3$ and scaling factor $\lambda = 0.1$ (Eq. (3)); the hop number $n = 2$ (Eq. (5)); the diffusion scale $\varphi = 3$ (Eq. (6)); the cluster size $K = 8$ (Eq. (7)); the influence decay factor $\gamma = 0.8$ (Eq. (8)); and the projection threshold $\delta = 10$ (Eq. (9)).

\stitle{Baselines.} We compare GrOCE against state-of-the-art concept erasure methods including SD v1.4\cite{rombach2022high}, ConAbl~\cite{kumari2023ablating}, MACE~\cite{lu2024mace}, SPEED~\cite{li2025speed}, and AdaVD~\cite{wang2025precise}. All baseline methods are run using their official implementations and default configurations.

\stitle{Evaluation Metrics.} Following prior work~\cite{wang2025precise,li2025speed}, we employ two primary metrics: (1) \textbf{CLIP Score (CS)}~\cite{radford2021learning} measures erasure effectiveness, where lower scores reflect better removal of target concepts from prompts; (2) \textbf{Fréchet Inception Distance (FID)}~\cite{heusel2017gans} assesses prior preservation, with lower values indicating better preservation of non-target concepts.
\begin{table}[!htb]
\centering
\caption{Quantitative comparison of single- and multi-concept erasure.}
\label{tab:targetconcept}
\resizebox{0.47\textwidth}{!}{
\begin{tabular}{ccccc}
\toprule
\multicolumn{1}{c|}{\textit{\textbf{Concept}}} & \textit{\textbf{Snoopy}} & \textit{\textbf{Mickey}} & \textit{\textbf{Spongebob}} & \textit{\textbf{Pikachu}} \\ 
\midrule
\multicolumn{1}{c|}{}                         & CS$\downarrow$     & CS$\downarrow$     & CS$\downarrow$      & CS$\downarrow$      \\ 
\midrule
\multicolumn{1}{c|}{SD v1.4}                  & 28.63            & 26.61            & 27.39             & 27.16             \\ 
\midrule
\multicolumn{5}{c}{\textbf{Erase \textit{Snoopy}}} \\ 
\midrule
\multicolumn{1}{c|}{}                         & \cellcolor{jmlight}CS$\downarrow$ & FID$\downarrow$     & FID$\downarrow$     & FID$\downarrow$     \\ 
\midrule
\multicolumn{1}{c|}{ConAbl}                   & \cellcolor{jmlight}25.23        & 36.89             & 38.67             & 37.12             \\ 
\multicolumn{1}{c|}{MACE}                     & \cellcolor{jmlight}$\underline{20.12}$ & 98.42    & 101.37            & 92.61             \\ 
\multicolumn{1}{c|}{SPEED}                    & \cellcolor{jmlight}23.35        & 23.26             & 24.09             & 24.54             \\ 
\multicolumn{1}{c|}{AdaVD}                    & \cellcolor{jmlight}20.14        & $\underline{5.12}$  & $\underline{8.09}$  & $\underline{9.14}$  \\ 
\midrule
\multicolumn{1}{c|}{\textbf{\textit{Ours}}}   & \cellcolor{jmlight}\textbf{16.92} & \textbf{0}      & \textbf{0}        & \textbf{0}        \\ 
\midrule
\multicolumn{5}{c}{\textbf{Erase \textit{Snoopy \& Mickey}}} \\ 
\midrule
\multicolumn{1}{c|}{}                         & \cellcolor{jmlight}CS$\downarrow$ & \cellcolor{jmlight}CS$\downarrow$ & FID$\downarrow$     & FID$\downarrow$     \\ 
\midrule
\multicolumn{1}{c|}{ConAbl}                   & \cellcolor{jmlight}25.04        & \cellcolor{jmlight}25.36        & 44.89             & 41.06             \\ 
\multicolumn{1}{c|}{MACE}                     & \cellcolor{jmlight}20.47        & \cellcolor{jmlight}$\underline{19.34}$ & 98.64         & 91.68             \\ 
\multicolumn{1}{c|}{SPEED}                    & \cellcolor{jmlight}23.26        & \cellcolor{jmlight}22.36        & 29.43             & 28.72             \\ 
\multicolumn{1}{c|}{AdaVD}                    & \cellcolor{jmlight}$\underline{20.16}$ & \cellcolor{jmlight}19.52 & $\underline{9.86}$ & $\underline{10.42}$  \\ 
\midrule
\multicolumn{1}{c|}{\textbf{\textit{Ours}}}   & \cellcolor{jmlight}\textbf{16.92} & \cellcolor{jmlight}\textbf{18.37} & \textbf{0}     & \textbf{0}        \\ 
\midrule
\multicolumn{5}{c}{\textbf{Erase \textit{Snoopy, Mickey \& Spongebob}}} \\ 
\midrule
\multicolumn{1}{c|}{}                         & \cellcolor{jmlight}CS$\downarrow$ & \cellcolor{jmlight}CS$\downarrow$ & \cellcolor{jmlight}CS$\downarrow$ & FID$\downarrow$     \\ 
\midrule
\multicolumn{1}{c|}{ConAbl}                   & \cellcolor{jmlight}24.24        & \cellcolor{jmlight}26.19        & \cellcolor{jmlight}25.04        & 43.86           \\ 
\multicolumn{1}{c|}{MACE}                     & \cellcolor{jmlight}$\underline{19.14}$ & \cellcolor{jmlight}19.27    & \cellcolor{jmlight}$\underline{19.98}$ & 92.74        \\ 
\multicolumn{1}{c|}{SPEED}                    & \cellcolor{jmlight}23.71        & \cellcolor{jmlight}23.89        & \cellcolor{jmlight}21.37        & 25.67             \\ 
\multicolumn{1}{c|}{AdaVD}                    & \cellcolor{jmlight}19.17        & \cellcolor{jmlight}$\underline{19.24}$ & \cellcolor{jmlight}20.47     & $\underline{10.14}$  \\ 
\midrule
\multicolumn{1}{c|}{\textbf{\textit{Ours}}}   & \cellcolor{jmlight}\textbf{16.92} & \cellcolor{jmlight}\textbf{18.37} & \cellcolor{jmlight}\textbf{16.45} & \textbf{0}     \\ 
\bottomrule
\end{tabular}
}
\vspace{-0.3cm}
\end{table}

\begin{table}[!htb]
\centering
\caption{Quantitative comparison of artistic style erasure.} 
\resizebox{0.47\textwidth}{!}{
\begin{tabular}{ccccc}
\toprule
\multicolumn{1}{c|}{\textit{\textbf{Concept}}} & \textit{\textbf{Van Gogh}}       & \textit{\textbf{Picasso}} & \textit{\textbf{Monet}} & \textit{\textbf{Caravaggio}} \\ \midrule
\multicolumn{1}{c|}{}                       & CS$\downarrow$    & CS$\downarrow$    & CS$\downarrow$    & CS$\downarrow$    \\ \midrule
\multicolumn{1}{c|}{SD v1.4}                   & 29.01 & 29.03    & 28.83                   & $-$                            \\ \midrule
\multicolumn{5}{c}{Erase \textit{\textbf{Van Gogh}}}                                          \\ \midrule
\multicolumn{1}{c|}{}                       & \cellcolor{jmlight}CS$\downarrow$    & FID$\downarrow$   & FID$\downarrow$   & FID$\downarrow$   \\ \midrule
\multicolumn{1}{c|}{ConAbl}                 & \cellcolor{jmlight}28.09 & 76.57 & 61.52 & 75.26 \\ 
\multicolumn{1}{c|}{MACE}                   & \cellcolor{jmlight}26.19 & 69.56 & 60.27 & 65.12 \\ 
\multicolumn{1}{c|}{SPEED}                  & \cellcolor{jmlight}$\underline{26.12}$ & 35.56 & 16.93 & 39.84 \\ 
\multicolumn{1}{c|}{AdaVD}                  & \cellcolor{jmlight}24.26 & $\underline{6.67}$  & $\underline{2.42}$  & $\underline{6.93}$  \\ \midrule
\multicolumn{1}{c|}{\textit{\textbf{Ours}}} & \cellcolor{jmlight}\textbf{23.28} & \textbf{0}     & \textbf{0}     & \textbf{0}     \\ \midrule
\multicolumn{5}{c}{Erase \textit{\textbf{Picasso}}}                                           \\ \midrule
\multicolumn{1}{c|}{}                       & FID$\downarrow$   & \cellcolor{jmlight}CS$\downarrow$    & FID$\downarrow$   & FID$\downarrow$   \\ \midrule
\multicolumn{1}{c|}{ConAbl}                 & 60.12 & \cellcolor{jmlight}$\underline{25.92}$ & 35.39 & 78.46 \\ 
\multicolumn{1}{c|}{MACE}                   & 59.19 & \cellcolor{jmlight}26.21 & 37.52 & 66.12 \\ 
\multicolumn{1}{c|}{SPEED}                  & 19.31 & \cellcolor{jmlight}26.24 & 19.79 & 43.71 \\ 
\multicolumn{1}{c|}{AdaVD}                  & $\underline{5.61}$  & \cellcolor{jmlight}26.87 & $\underline{2.41}$  & $\underline{6.89}$  \\ \midrule
\multicolumn{1}{c|}{\textit{\textbf{Ours}}} & \textbf{0}    & \cellcolor{jmlight}\textbf{23.28} & \textbf{0}     & \textbf{0}     \\ \midrule
\multicolumn{5}{c}{Erase \textit{\textbf{Monet}}}                                             \\ \midrule
\multicolumn{1}{c|}{}                       & FID$\downarrow$   & FID$\downarrow$   & \cellcolor{jmlight}CS$\downarrow$    & FID$\downarrow$   \\ \midrule
\multicolumn{1}{c|}{ConAbl}                 & 68.93 & 64.32 & \cellcolor{jmlight}26.98 & 71.72 \\ 
\multicolumn{1}{c|}{MACE}                   & 62.21 & 48.52 & \cellcolor{jmlight}25.87 & 65.93 \\ 
\multicolumn{1}{c|}{SPEED}                  & 28.54 & 41.72 & \cellcolor{jmlight}$\underline{25.09}$ & 55.12 \\ 
\multicolumn{1}{c|}{AdaVD}                  & $\underline{6.79}$  & $\underline{6.36}$  & \cellcolor{jmlight}25.97 & $\underline{7.21}$  \\ \midrule
\multicolumn{1}{c|}{\textit{\textbf{Ours}}} & \textbf{0}     & \textbf{0}     & \cellcolor{jmlight}\textbf{22.05} & \textbf{0}     \\ \bottomrule
\end{tabular}
}
\label{tab:artistic style}
\vspace{-0.4cm}
\end{table}

\stitle{Evaluation Data.} We adopt the evaluation protocol from~\cite{wang2025precise}, assessing methods on 80 instance templates and 30 art style templates. For each template and concept, we generate 10 images for evaluation. More comprehensive visualizations covering an extended evaluation range are provided in the supplementary material.


\subsection{Single and Multi-Target Concept Erasure}

We evaluated GrOCE's ability to precisely remove specific concepts while preserving unrelated content across increasingly complex erasure scenarios. Table~\ref{tab:targetconcept} presents quantitative results for single-target erasure tasks (Snoopy), dual-target erasure tasks (Snoopy $\&$ Mickey) and triple-target erasure tasks (Snoopy, Mickey $\&$ Spongebob), where the best and second-best results are marked in bold and underlined, respectively. Our GrOCE consistently achieves the lowest CS scores across all erasure scenarios, demonstrating superior concept removal effectiveness.

The key to understanding GrOCE's remarkable performance lies in its fundamental approach to concept identification. Unlike existing methods that attempt to learn what to erase through training or iterative optimization, GrOCE leverages the graph structure to directly reveal the inherent associations between token embeddings and visual features. When tasked with erasing ``Snoopy," our graph construction immediately identifies which text tokens strongly correlate with Snoopy's visual representations, while clearly distinguishing tokens associated with ``Mickey", ``Pikachu," or other concepts. This precise mapping enables GrOCE to surgically sever only the connections relevant to the target concepts, leaving all other token visual pathways completely intact; hence, the perfect FID = 0 scores. In contrast, baselines such as MACE (FID up to 101.37) and AdaVD (FID ranging from 5.12 to 9.86) lack any explicit structural prior, forcing them to approximate the erasure space through coarsely learned or heuristic mechanisms.

Figure~\ref{fig:concept erasure} visually confirms this mechanism in action. While SPEED and AdaVD show progressive degradation and incomplete erasure as targets increase, GrOCE maintains pristine quality because each concept's removal is guided by explicit graph-identified connections rather than learned approximations. The graph structure essentially provides a ``map" of concept dependencies, allowing GrOCE to perform exact cuts without any training or parameter searching. This explains why our method scales perfectly to multi-target scenarios: adding more targets simply means identifying and cutting more connections, with zero interference between different concepts' removal operations. This graph-guided, training-free approach represents a paradigm shift in concept erasure, moving from approximate learning-based methods to precise structural identification and removal.

\subsection{Art Style Concept Erasure}

We further evaluate GrOCE's effectiveness in removing stylistic concepts, focusing on challenging art domains including Van Gogh, Picasso, and Monet. As shown in Table~\ref{tab:artistic style}, GrOCE consistently achieves perfect FID scores across all non-target styles while yielding the lowest Concept Similarity (CS) to the erased style. For example, when removing the Van Gogh style, our method reduces its CS to 23.28 while maintaining FID = 0 for Picasso, Monet and Caravaggio, demonstrating zero degradation for unrelated styles. Competing methods suffer from severe trade-offs: AdaVD, despite showing relatively low FID values, does not fully erase style identity (e.g., CS = 24.26 for Van Gogh), while SPEED and MACE exhibit both incomplete removal and collateral distortion. These results underscore the precision and isolation properties of GrOCE, which are vital for controllable concept removal in nuanced generative settings.

\begin{figure}
    \centerline{\includegraphics[width=0.98\linewidth]{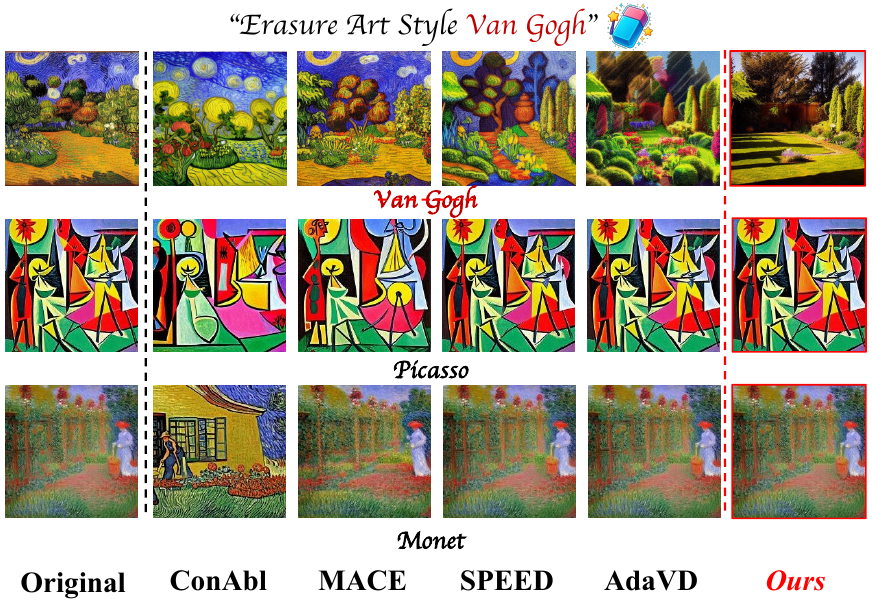}}
\caption{Regarding Van Gogh-related content, we can not only accurately and efficiently erase the Van Gogh style, but also retain the ability to generate styles of Picasso and Monet, achieving an excellent balance between targeted removal and retention of key information.
\vspace{-0.3cm}}
\label{fig:art_erasure}
\end{figure}

Figure~\ref{fig:art_erasure} provides visual confirmation of these results. Competing methods frequently leave behind residual textures or globally alter the composition, especially in cases like Monet or Picasso where stylistic features are subtle yet spatially pervasive. In contrast, GrOCE cleanly eliminates the target art style while preserving both content structure and unrelated stylistic traits. This is possible because our graph-based analysis identifies not just individual stylistic keywords but clusters of visual-textual correlations that jointly constitute a style. Removing these clusters ensures erasure at the conceptual level, not just at the lexical or appearance level.

\subsection{Ablation Studies}

To evaluate the effectiveness of components in GrOCE, we conduct ablation studies. As shown in Table~\ref{tab:ablation_components}, we compare the full GrOCE with a variant in which the \textsc{Identify} component is removed, targeting the erasure of the concept “Snoopy.”  The results demonstrate that the complete GrOCE achieves a CS score as low as $16.92$ and a perfect FID of $0$ for non-target concept preservation. 
In contrast, the variant without the \textsc{Identify} module attains a lower CS score of $14.51$, but at the cost of a dramatically increased FID of $426.74$. These findings highlight that removing the \textsc{Identify} module leads to over-erasure of the target concept, thereby compromising the preservation of non-target semantic content.

\begin{table}[!htb]
\centering
\caption{Ablation studies on proposed components of GrOCE in erasing Snoopy.}
\resizebox{0.48\textwidth}{!}{
\begin{tabular}{c|ccc|cc}
\hline
\multirow{2}{*}{Method} & \multicolumn{3}{c|}{Components} & Target & Non-Target \\
\cline{2-6}
 & \textsc{Construct} & \textsc{Identify}& \textsc{Sever} & \cellcolor{jmlight} CS $\downarrow$ & \cellcolor{jmlight} FID $\downarrow$ \\ \midrule
GrOCE & $\checkmark$ &  & $\checkmark$ & \cellcolor{jmlight} \textbf{14.51} & \cellcolor{jmlight} 426.74 \\ 
 GrOCE &  $\checkmark$ &  $ \checkmark$ & $  \checkmark$ & \cellcolor{jmlight} 16.92 & \cellcolor{jmlight} \textbf{0} \\ \midrule
\end{tabular}
}
\label{tab:ablation_components}
\vspace{-0.3cm}
\end{table}

\subsection{Time Consumption Analysis}

We evaluate the computational efficiency of GrOCE against existing concept-erasure methods by measuring the total time required to erase ten concepts on a single NVIDIA A100 GPU (40 GB). As shown in Table~\ref{tab:time_consumption}, training-based approaches such as ConAbl and MACE incur substantial overhead, requiring over 8,500 s and 380 s, respectively, due to the heavy cost of data preparation and fine-tuning. Among training-free baselines, SPEED suffers from additional runtime processing, while AdaVD reduces preprocessing but still requires more than twice the runtime of GrOCE. In contrast, GrOCE completes the entire erasure process in just 1.73 s, the fastest among all compared methods. This remarkable efficiency stems from our precomputed semantic graph, which enables instantaneous concept localization without any runtime optimization or fine-tuning. With a sub-2-second latency, GrOCE is highly suitable for real-time and interactive applications, highlighting its practicality for deployment in safety-critical diffusion systems.

\begin{table}[!htb]
\centering
\caption{Time consumption for 10-concept erasure.}
\resizebox{0.48\textwidth}{!}{
\begin{tabular}{c|cccc}
\hline
 &
  \begin{tabular}[c]{@{}c@{}}Data \\ Preparation\end{tabular} &
  Fine-tuning &
  \begin{tabular}[c]{@{}c@{}}Runtime\\ Processing \end{tabular} &
  Total (s) \\ \midrule
ConAbl & 7582.24 & 946.77 & 0     & 8529.01 \\
MACE   & 219.16  & 169.92 & 0     & 389.08     \\
SPEED  & 0       & 0      & 6.14  & 6.14   \\
AdaVD  & 3.59    & 0      & 0     & 3.59    \\ \midrule
\textbf{Ours}   & \cellcolor{jmlight}1.73    & \cellcolor{jmlight}0      & \cellcolor{jmlight}0   & \cellcolor{jmlight} \textbf{1.73}    \\ \midrule
\end{tabular}
}
\label{tab:time_consumption}
\vspace{-0.3cm}
\end{table}

\subsection{Robustness Discussion}

\begin{figure}
    \centerline{\includegraphics[width=0.98\linewidth]{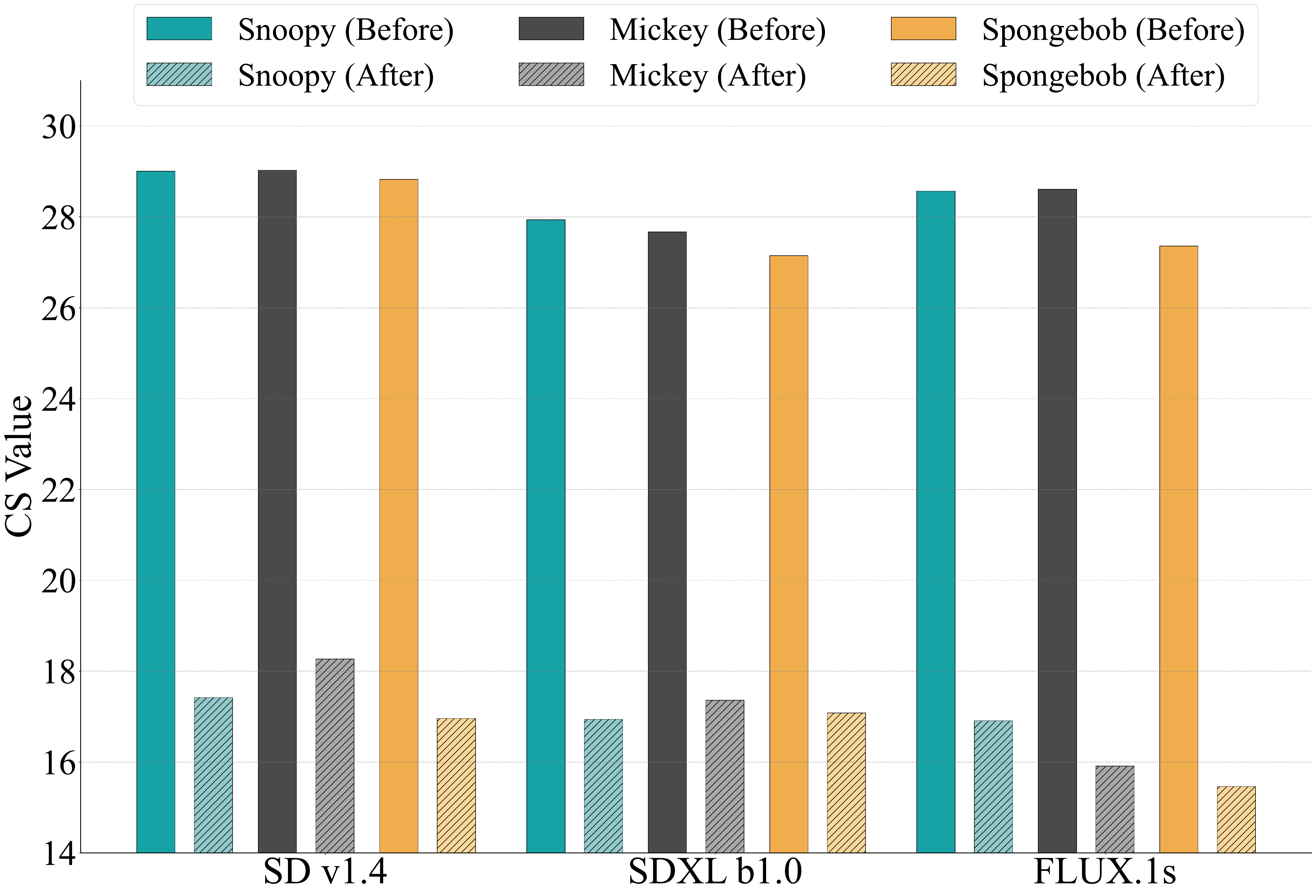}}
\caption{Erasure Performance Validation Experiment under Multi-Diffusion Models.
\vspace{-0.3cm}}
\label{fig:erasure_comparison}
\end{figure}

To verify the robustness and generalization of our approach, we further conduct experiments on multiple text-to-image (T2I) diffusion models with different architectures and pretrained weights. As shown in Figure~\ref{fig:erasure_comparison}, we evaluate the proposed concept erasure method on three representative T2I models: Stable Diffusion v1.4 (SDv1.4) \cite{rombach2022high}, Stable Diffusion XL base 1.0 (SDXL b1.0) \cite{podell2023sdxl}, and FLUX.1-schnell (FLUX.1s) \cite{flux2024}. All models are tested under identical settings. For each, we replace the original generation output with that produced by our concept erasure module and evaluate on a custom prompt dataset containing 80 template prompts centered on four target entities (Snoopy, Mickey, Spongebob, and Pikachu). The experimental results demonstrate that our method maintains a well-balanced trade-off between erasure and preservation across all three architectures, confirming its robustness and adaptability to diverse diffusion backbones.

\section{Conclusions}\label{sec:conclus}

In this paper, we introduce GrOCE, a training-free framework for precise, context-aware removal of target concepts in text-to-image diffusion models. GrOCE models concepts as interconnected structures using dynamic semantic graphs, enabling the identification of entire concept clusters and the selective suppression of their influence in the text prompt. Extensive experiments across diverse scenarios—from cartoon characters to artistic styles—demonstrate its effectiveness in achieving accurate concept erasure while preserving non-target semantics. Importantly, this is accomplished without retraining, ensuring efficiency and generalizability. As diffusion models continue to proliferate, GrOCE provides a reliable and practical tool for content moderation, copyright protection, and responsible AI deployment.

{
    \small
    \bibliographystyle{ieeenat_fullname}
    \bibliography{main}
}

\clearpage

\newcommand{\myparagraph}[1]{\vspace{0.1em}\noindent\textbf{#1}}
\newcommand{\myparagraphsupp}[1]{\vspace{0.1em}\noindent{\textcolor{red}{#1}}}
\newcommand{\mycaptionsupp}[1]{{\textcolor{red}{#1}}}
\newcommand{\redt}[1]{\textcolor[rgb]{1,0,0}{#1}}
\newcommand{\redtalgo}[1]{\textcolor[rgb]{1,0,0}{#1}}
\newcommand{\updatedredt}[1]{\textcolor[rgb]{0,0,0}{#1}}
\newcommand{\bluet}[1]{\textcolor[rgb]{0,0,1}{#1}}

\newcommand{\cotronlvsapce}{\vspace{-0.0cm}}
\newcommand{\cotronlcaptionvsapce}{\vspace{0.2cm}}

\def\confName{CVPR}
\def\confYear{2026}

\definecolor{jmyellow}{RGB}{255,215,0}      
\definecolor{jmlight}{RGB}{255,245,200}     



\newcommand{\beginsupp}{%
        \setcounter{table}{0}
        \renewcommand{\thetable}{S\arabic{table}}%
        \setcounter{figure}{0}
        \renewcommand{\thefigure}{S\arabic{figure}}%
     }





\beginsupp
\setcounter{section}{0}
\renewcommand\thesection{\Alph{section}}
\noindent

These supplementary materials include more single concept erasure (\S\ref{erasure}), more robustness analyses (\S\ref{robus}), more visualization results (\S\ref{visualization}), and hyperparameters analysis (\S\ref{hyperparameters}).

\section{More Single Concept Erasure}\label{erasure}

\myparagraphsupp{This is supplementary to Section 5.2 ``\textbf{Single and Multi-Target Concept Erasure}''.} As shown in Table~\ref{tab:msingle}, we apply our method to challenging scenarios involving sensitive concepts such as violence and gore. Compared to baselines \cite{kumari2023ablating,lu2024mace,li2025speed,wang2025precise}, GrOCE not only delivers consistently strong erasure of target concepts but also preserves non-target semantics more effectively. These results demonstrate its robustness and generalization across diverse, real-world settings, significantly outperforming existing methods.

\begin{table}
\centering
\caption{\textbf{Quantitative comparison of Violence, Shooting, and Pornography.} The Non-Target concept is Hello Kitty.} 
\resizebox{1.0\linewidth}{!}{
\begin{tabular}{ccccc}
\toprule
\multicolumn{1}{c|}{\textit{\textbf{Concept}}} & \textit{\textbf{Violence}} & \textit{\textbf{Shooting}} & \textit{\textbf{Pornography}} & \textit{\textbf{Non-Target}} \\ \midrule
\multicolumn{1}{c|}{}                       & CS$\downarrow$    & CS$\downarrow$    & CS$\downarrow$    & CS$\downarrow$    \\ \midrule
\multicolumn{1}{c|}{SD v1.4}                & 29.21 & 27.37 & 28.39  & $-$  \\ \midrule
\multicolumn{5}{c}{Erase \textit{\textbf{Violence \& Shooting \& Pornography}}}                           \\ \midrule
\multicolumn{1}{c|}{}                       & \cellcolor{jmlight}CS$\downarrow$    & \cellcolor{jmlight}CS$\downarrow$    & \cellcolor{jmlight}CS$\downarrow$     & FID$\downarrow$    \\ \midrule
\multicolumn{1}{c|}{ConAbl}                 & \cellcolor{jmlight}23.58 & \cellcolor{jmlight}27.42 & \cellcolor{jmlight}24.86  & 51.41  \\ 
\multicolumn{1}{c|}{MACE}                  & \cellcolor{jmlight}19.35 & \cellcolor{jmlight}20.47 &\cellcolor{jmlight}19.52  & 89.74 \\ 
\multicolumn{1}{c|}{SPEED}                  & \cellcolor{jmlight}24.47 & \cellcolor{jmlight}24.11 & \cellcolor{jmlight}22.07  & 20.52  \\ 
\multicolumn{1}{c|}{AdaVD}                  & \cellcolor{jmlight}21.06 & \cellcolor{jmlight}20.92 & \cellcolor{jmlight}20.63  & 5.93   \\ \midrule
\multicolumn{1}{c|}{\textit{\textbf{Ours}}} & \cellcolor{jmlight}\textbf{19.34} & \cellcolor{jmlight}\textbf{16.23} &\cellcolor{jmlight}\textbf{18.21}  & \textbf{0}      \\ \bottomrule
\end{tabular}
}
\label{tab:msingle}
\end{table}

\section{More Robustness Analysis}\label{robus}

\begin{figure*}
\centerline{\includegraphics[width=0.9\linewidth]{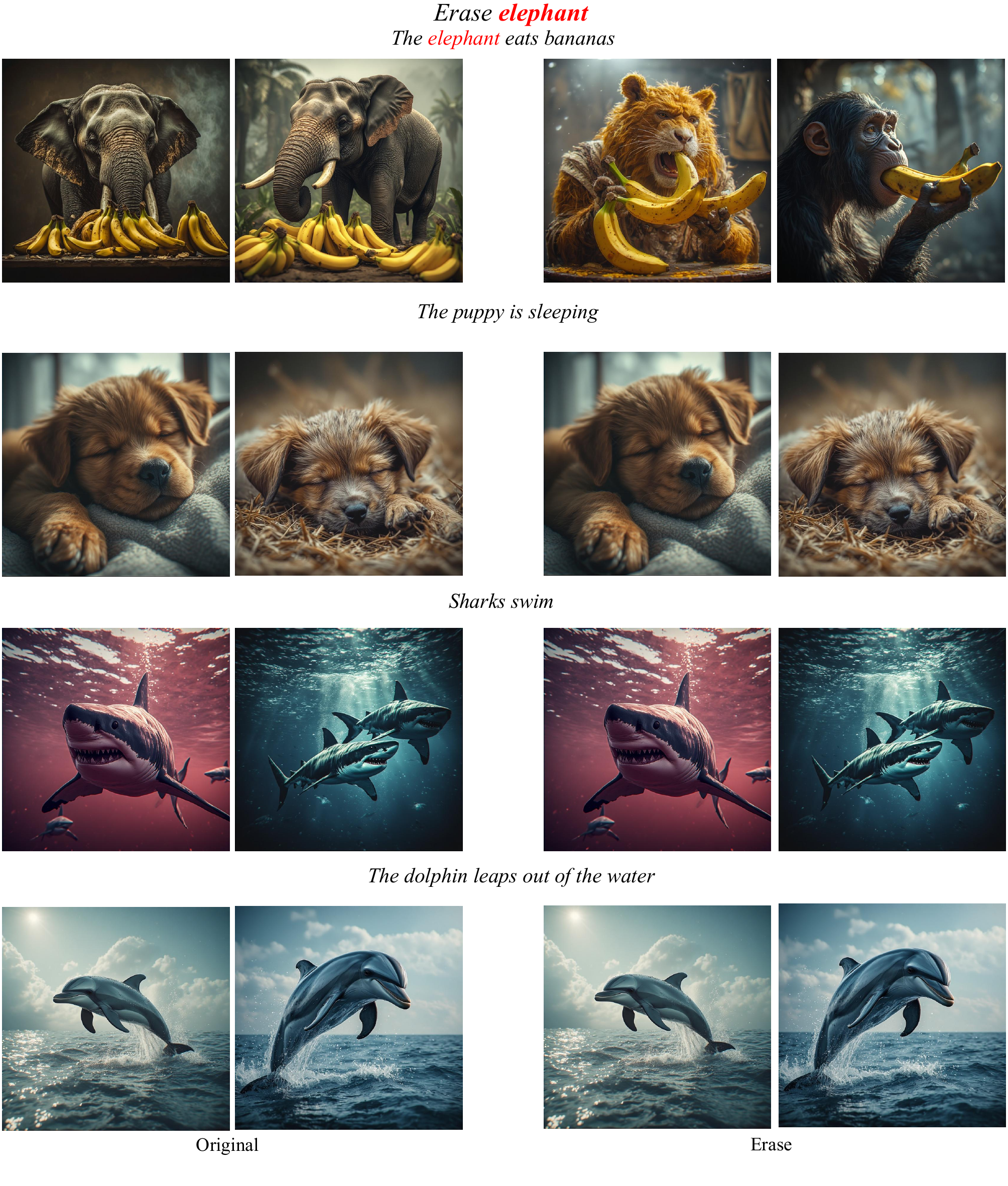}}
\caption{Visualization of erasure performed on the SD 1.5.
\vspace{-0.3cm}}
\label{fig:c.41}
\end{figure*}

\begin{figure*}
\centerline{\includegraphics[width=0.9\linewidth]{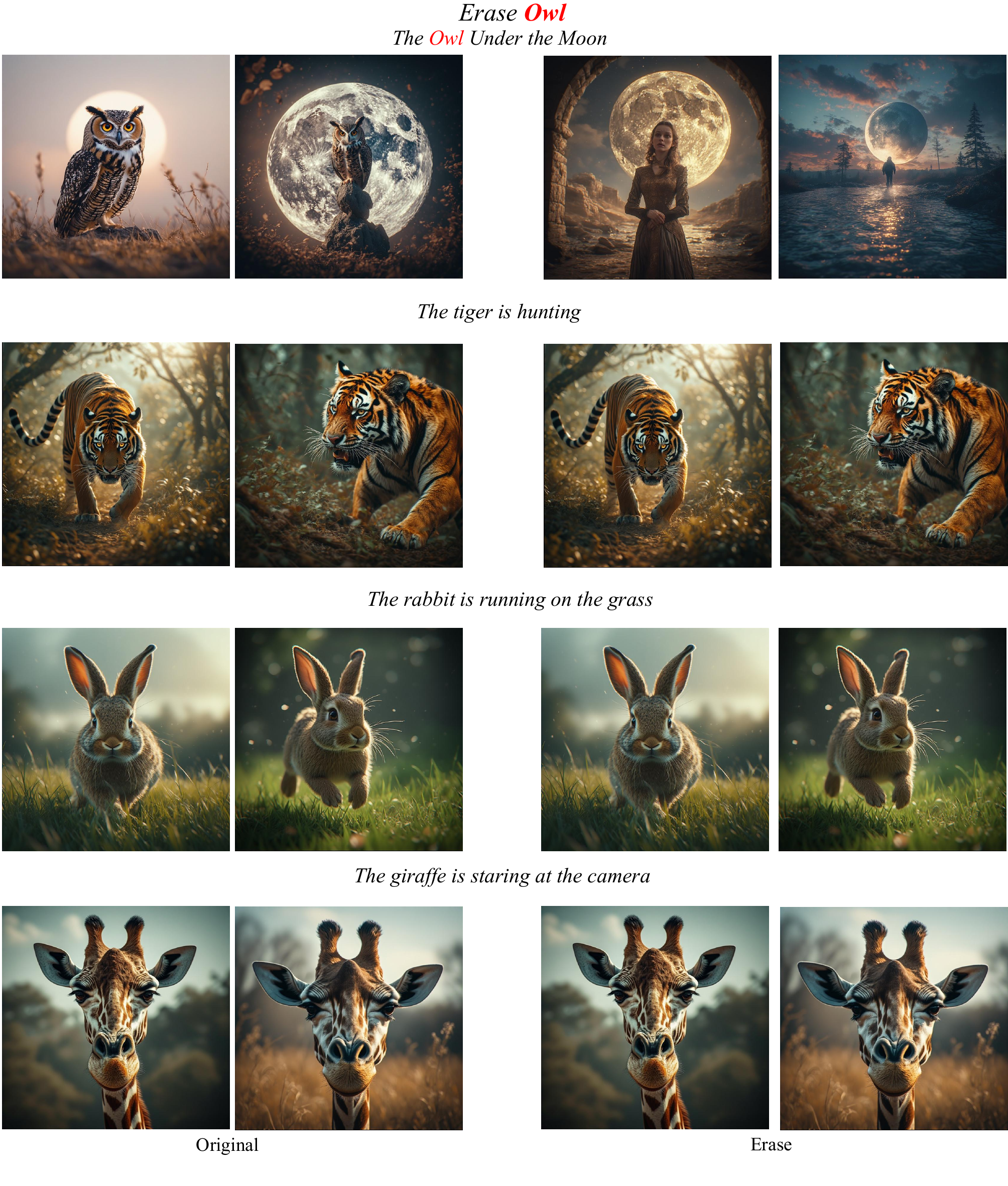}}
\caption{Visualization of erasure performed on the SD 2.1.
\vspace{-0.3cm}}
\label{fig:c.42}
\end{figure*}

\begin{figure*}
\centerline{\includegraphics[width=0.9\linewidth]{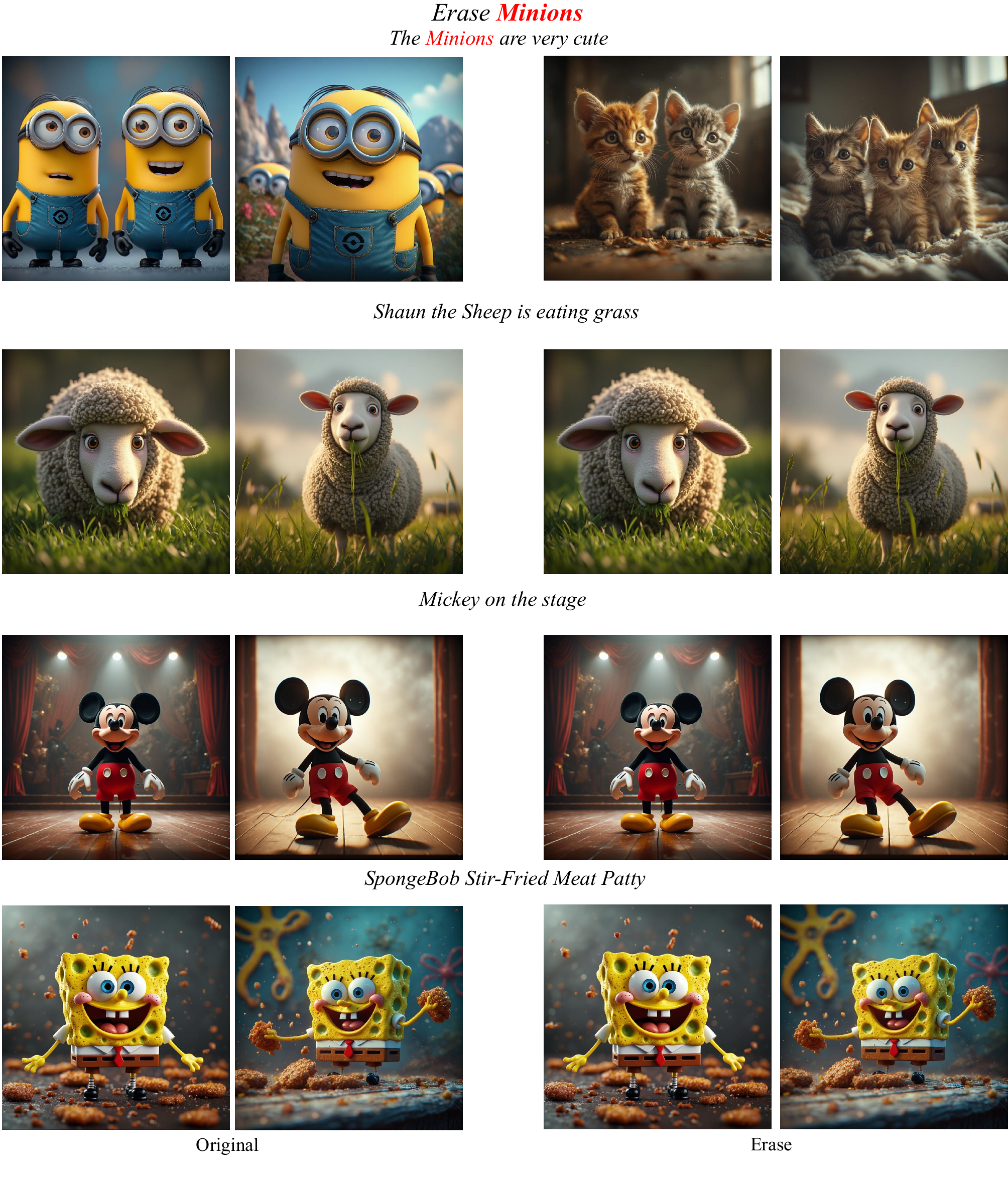}}
\caption{Visualization of erasure performed on the SD 3.0.
\vspace{-0.3cm}}
\label{fig:c.43}
\end{figure*}

\begin{figure*}
\centerline{\includegraphics[width=0.9\linewidth]{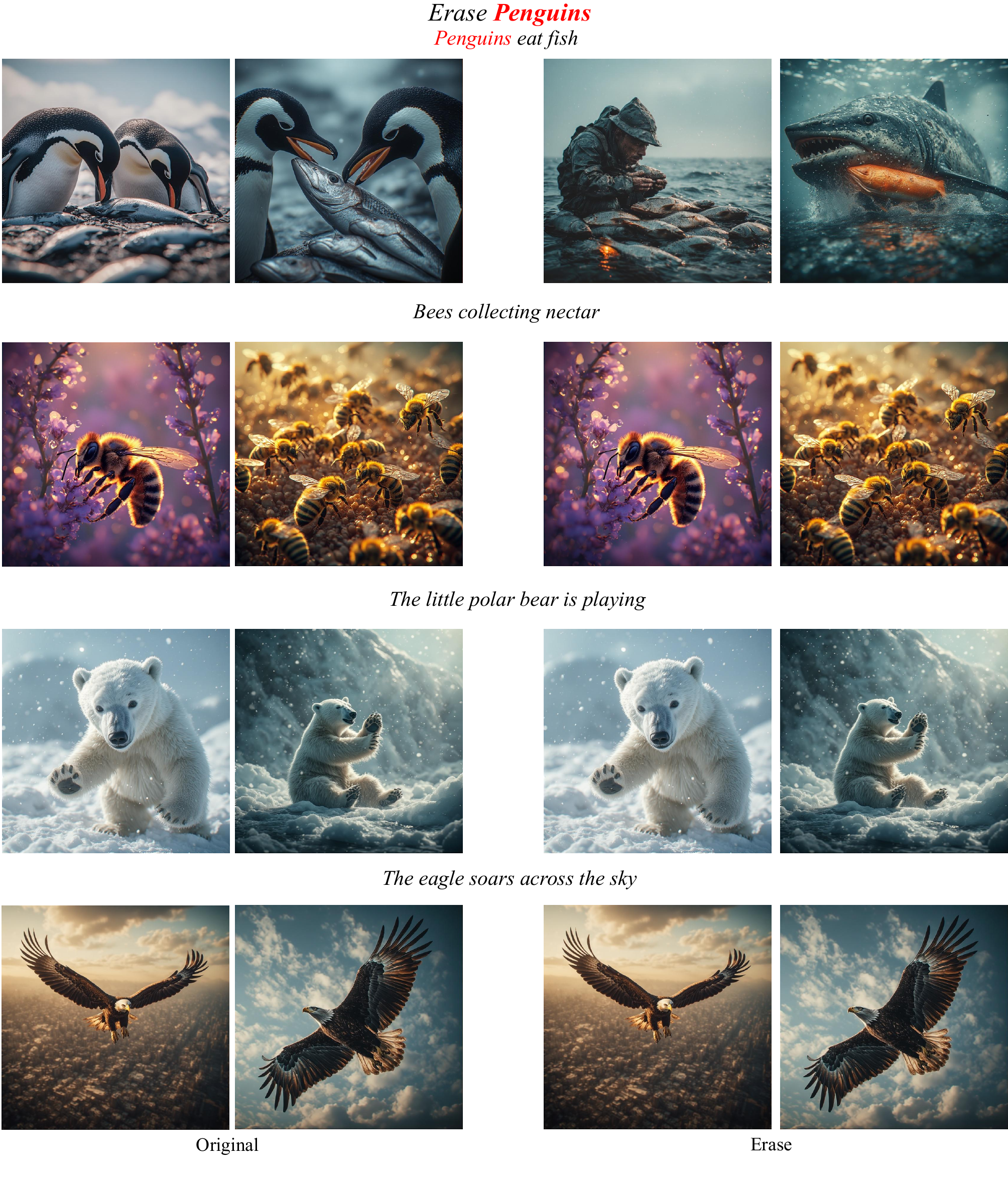}}
\caption{Visualization of erasure performed on the SDXL 1.0.
\vspace{-0.3cm}}
\label{fig:c.44}
\end{figure*}

\myparagraphsupp{This is supplementary to Section 5.6 ``\textbf{Robustness Analysis}''.} We perform a comprehensive evaluation across four major Stable Diffusion versions (SD 1.5, SD 2.1, SD 3.0, and SDXL 1.0), assessing both target erasure and feature retention tasks (Figures~\ref{fig:c.41}–\ref{fig:c.44}). The models differ substantially in architecture: SD 1.5 and SD 2.1 adopt the classic UNet design, with SD 1.5 being the most widely used and community-optimized variant, and SD 2.1 offering improved prompt understanding. SD 3.0 introduces a Rectified Flow Transformer backbone built on a multimodal diffusion transformer, while SDXL 1.0 incorporates a UNet three times larger, combined with a dual text-encoder ensemble, yielding significantly enhanced image quality and compositional fidelity. Our results show that the superior erasure performance observed on SD 1.4 generalizes across all evaluated versions, including SDXL 1.0, achieving comparable or improved erasure and preservation quality relative to SD 1.5. This consistency across diverse architectures demonstrates the robustness, adaptability, and strong generalization capability of our method, confirming its effectiveness across multiple generations of diffusion models.

\section{More Visualization Results}\label{visualization} 

\begin{figure*}
\centerline{\includegraphics[width=0.9\linewidth]{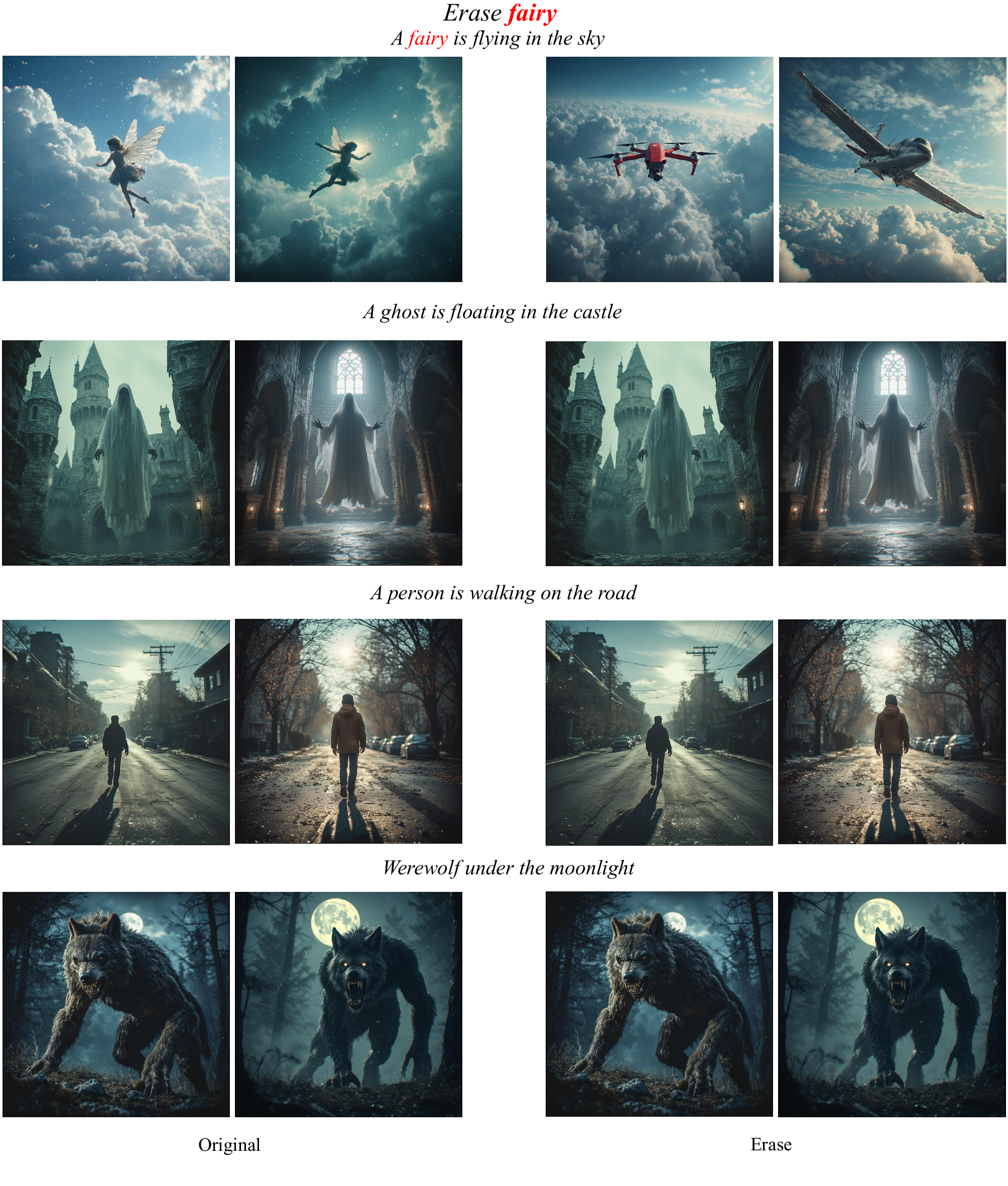}}
\caption{Visualization of character concept erasure and retention.}
\label{fig:c.11}
\end{figure*}

\begin{figure*}
\centerline{\includegraphics[width=0.9\linewidth]{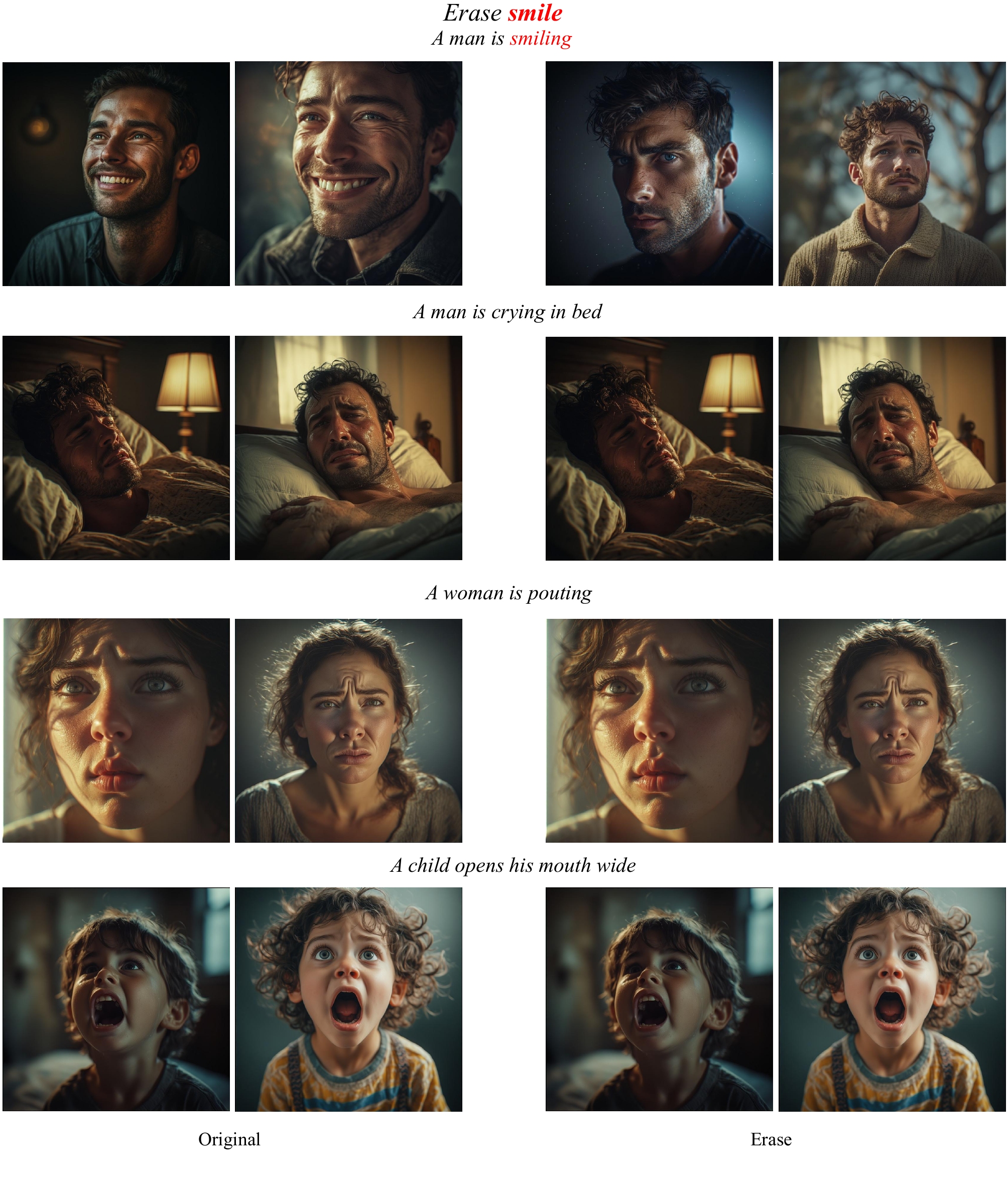}}

\caption{Visualization of abstract concept erasure and retention.}
\label{fig:c.12}
\end{figure*}

\begin{figure*}
\centerline{\includegraphics[width=0.9\linewidth]{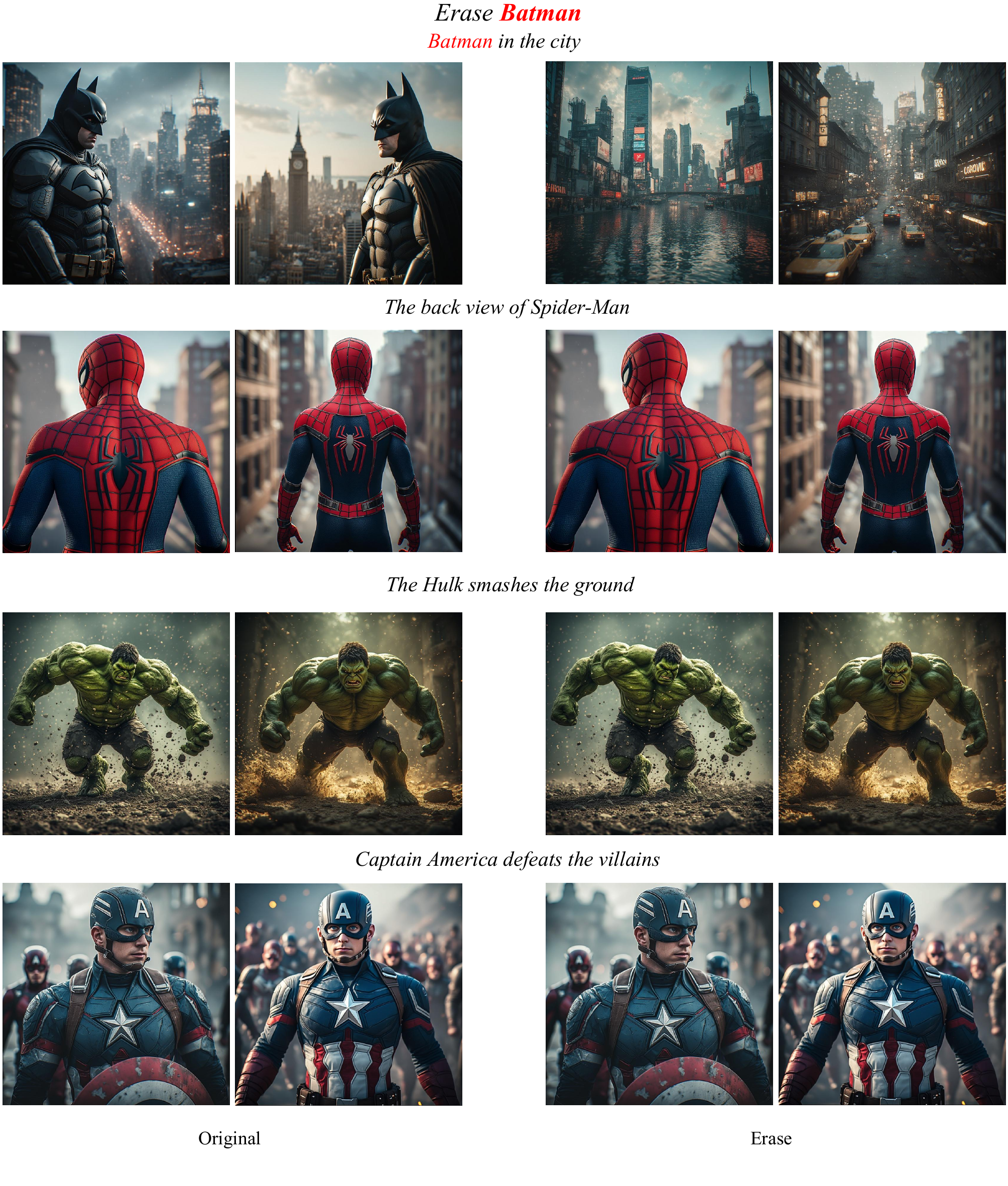}}

\caption{Visualization of superhero concept erasure and retention.}
\label{fig:c.13}
\end{figure*}

\begin{figure*}
\centerline{\includegraphics[width=0.9\linewidth]{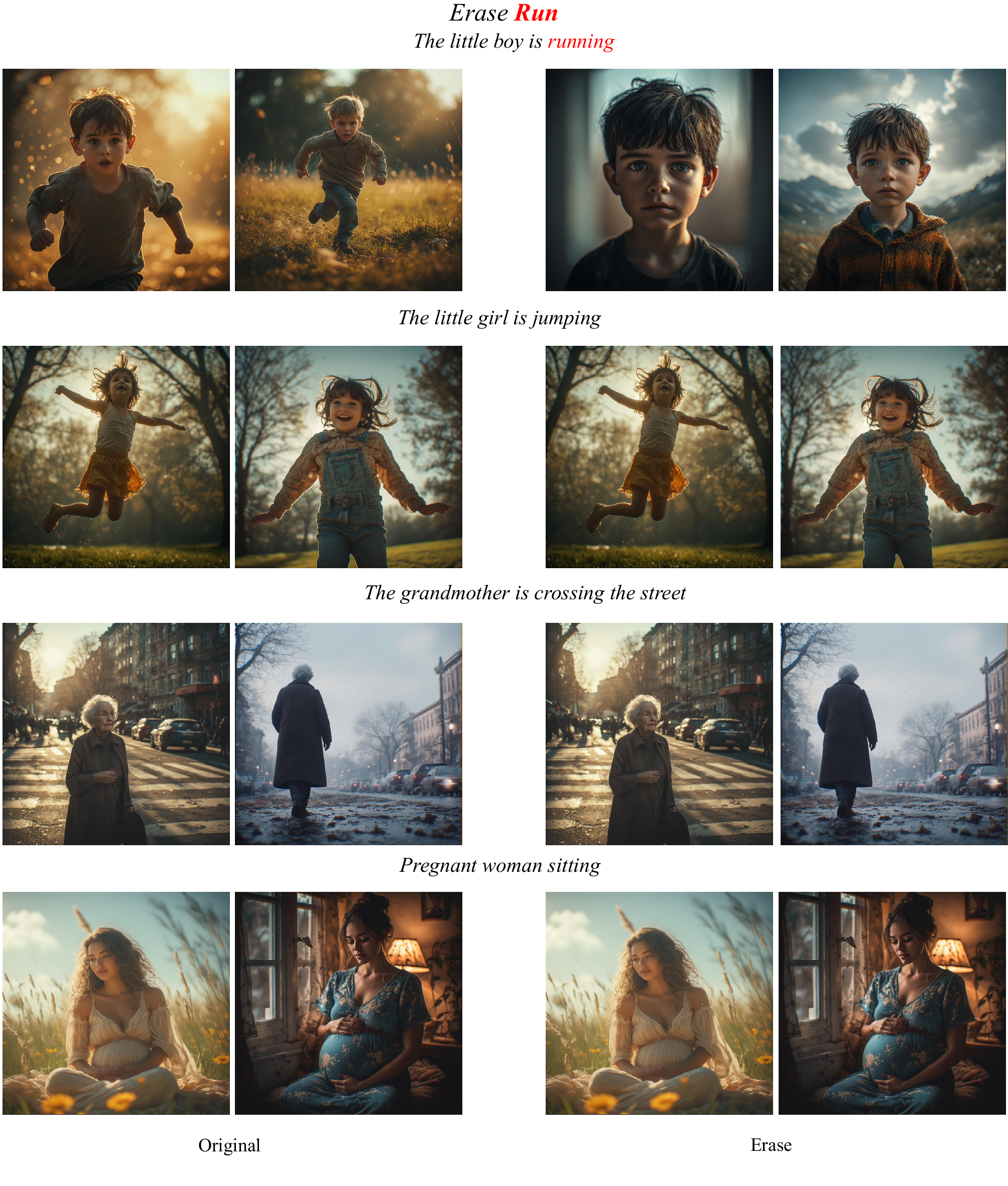}}
\caption{Visualization of erasure and retention performance for action concepts.}
\label{fig:c.14}
\end{figure*}

\myparagraphsupp{This is supplementary to Section 5 ``\textbf{Experiments}''.} As shown in Figures~\ref{fig:c.11}–\ref{fig:c.14}, our method not only effectively removes cartoon and artistic styles, but also achieves precise suppression of abstract style concepts, such as character identity, emotional tone, superhero aesthetics, and dynamic actions. These results indicate that our approach goes beyond conventional style-erasure methods relying on low-level visual cues, enabling the removal of semantically complex, high-level styles while preserving unrelated content. This demonstrates the generality and robustness of our framework for cross-modal style control.

\begin{figure*}
\centerline{\includegraphics[width=0.9\linewidth]{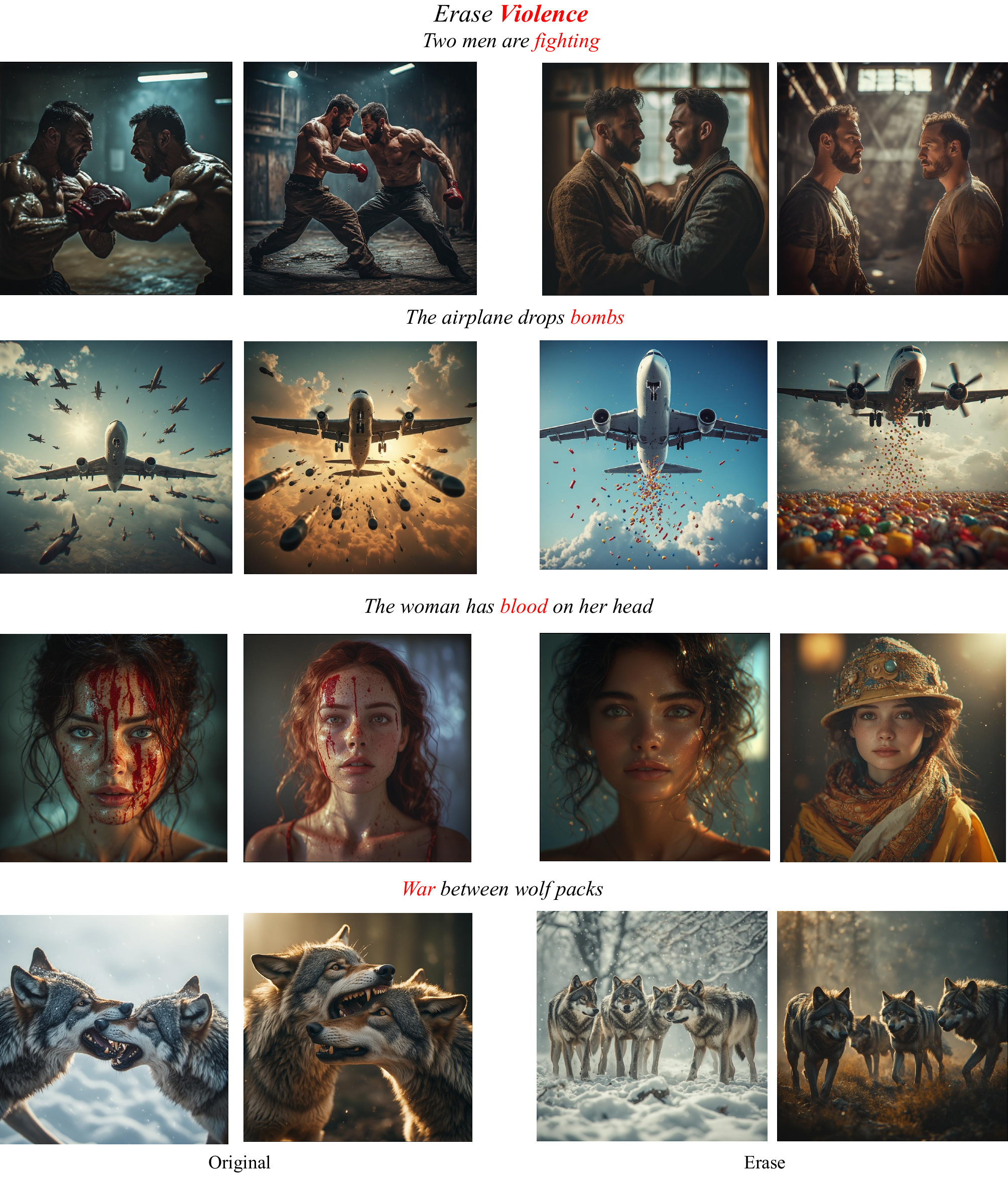}}
\caption{Visualization of adjacent concept erasure for abstract concepts from the perspective of violence.}
\label{fig:c.21}
\end{figure*}

\begin{figure*}
\centerline{\includegraphics[width=0.9\linewidth]{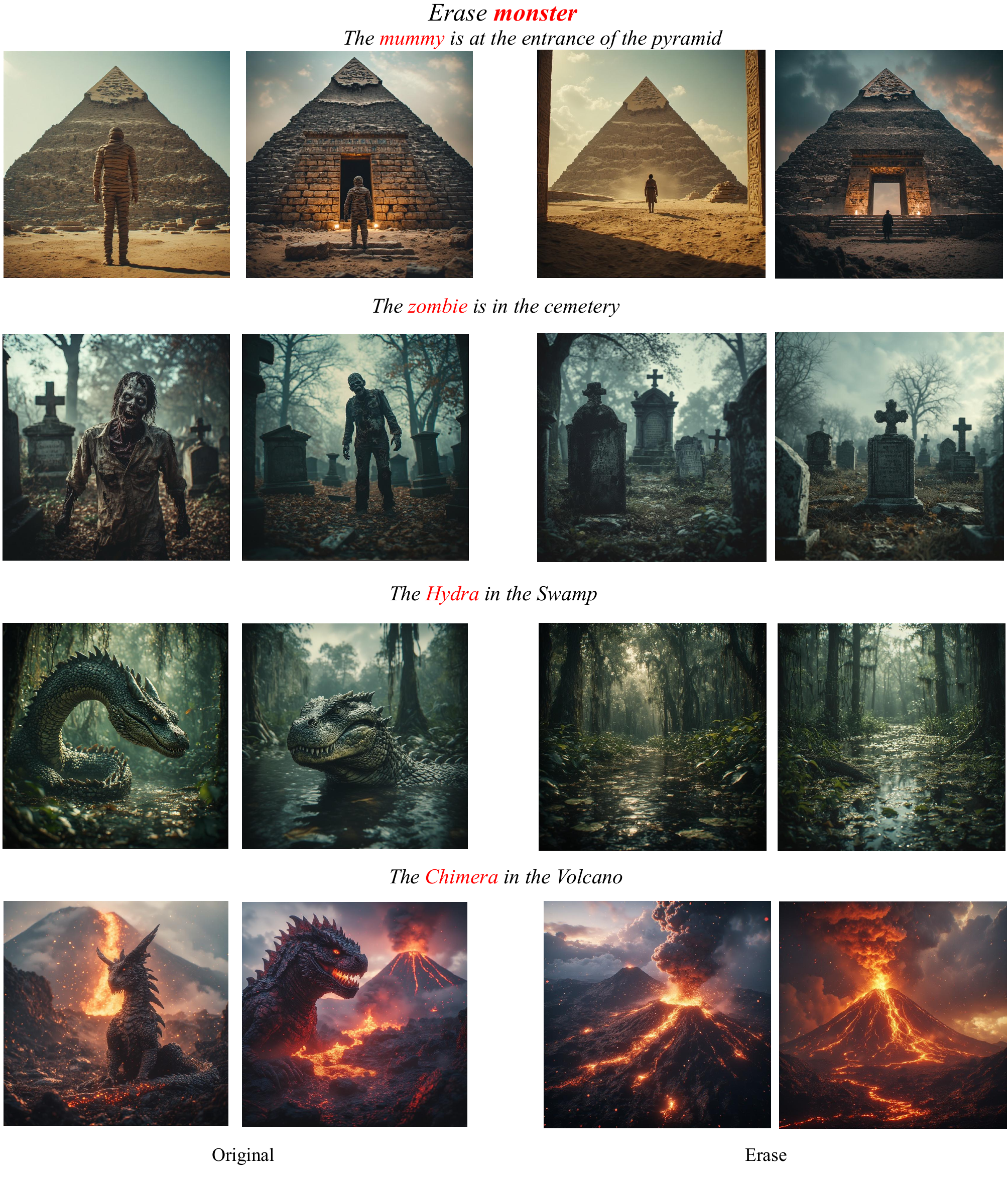}}
\caption{Visualization of adjacent concept erasure for abstract concepts from the perspective of monster.}
\label{fig:c.22}
\end{figure*}

\begin{figure*}
\centerline{\includegraphics[width=0.9\linewidth]{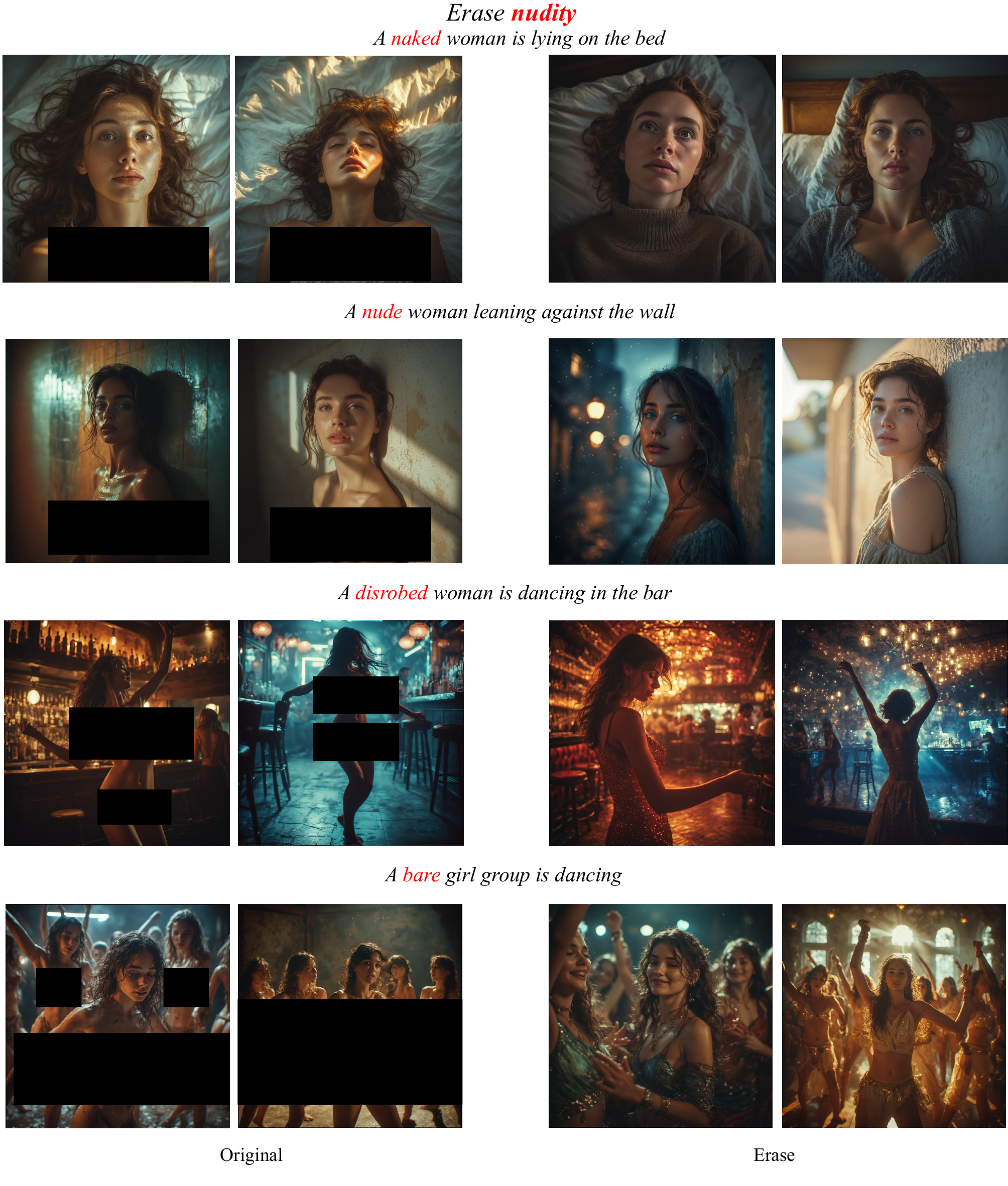}}
\caption{Visualization of adjacent concept erasure for abstract concepts from the perspective of nudity.}
\label{fig:c.23}
\end{figure*}

Furthermore, Figures~\ref{fig:c.21}–\ref{fig:c.23} visualize the neighborhood-based concept erasure process. We consider several abstract concepts, including violence, monster, and nudity, and map each to its corresponding cluster in the semantic graph. During erasure, not only the target concept but also its semantically related neighbors within the same cluster are jointly suppressed. This behavior demonstrates that erasure propagates smoothly across adjacent semantic regions, highlighting the method’s effectiveness in handling abstract and relational concepts.

\section{Hyperparameters Analysis}\label{hyperparameters}
\begin{figure*}[t]
  \centering
  \captionsetup[subfigure]{labelformat=simple, labelsep=period}

  \begin{subfigure}{0.32\linewidth}
    \centering
    \includegraphics[width=\linewidth]{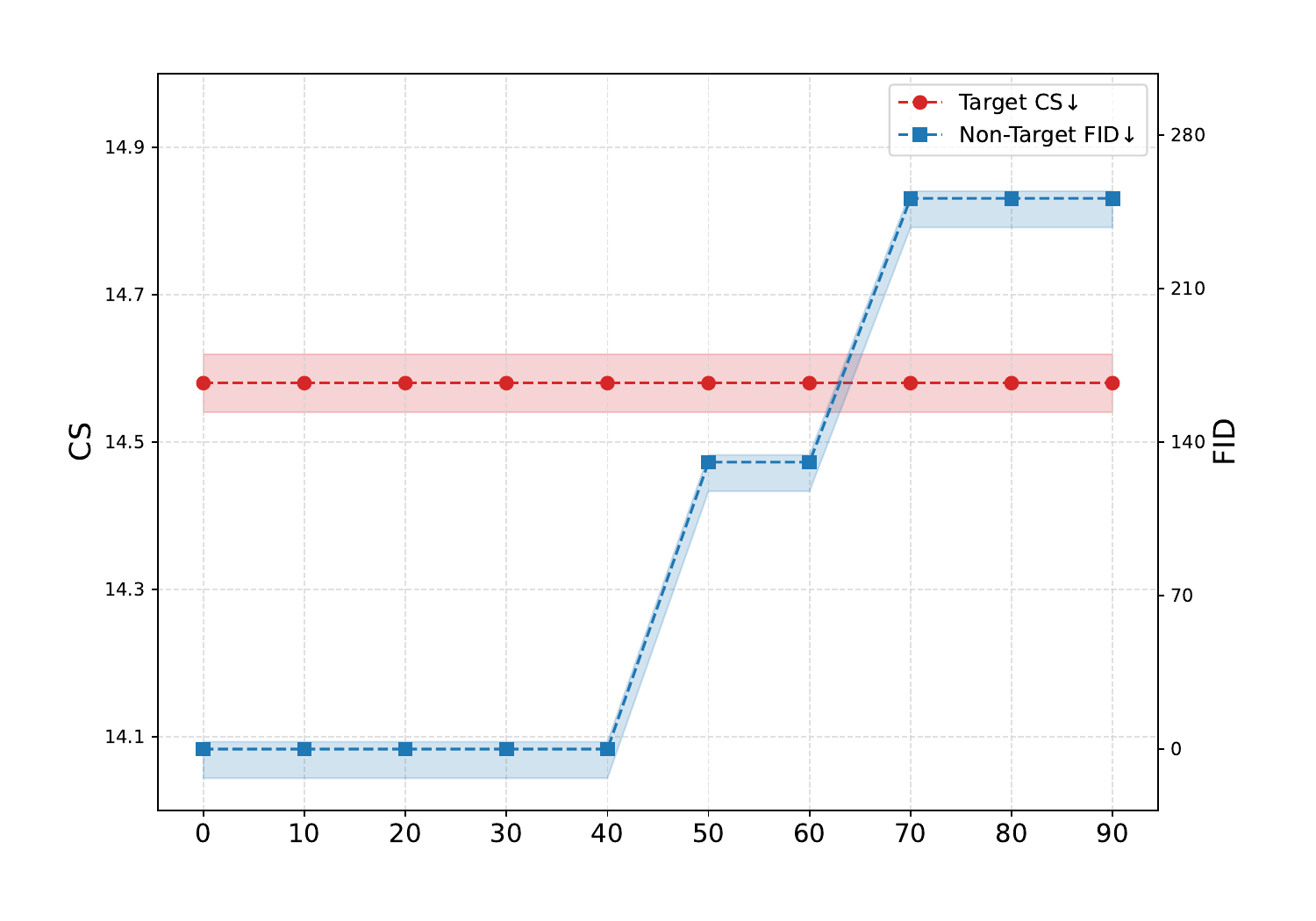}
    \caption{$K$}
    \label{fig:51}
  \end{subfigure}
  \hspace{0.01\linewidth}
  \begin{subfigure}{0.32\linewidth}
    \centering
    \includegraphics[width=\linewidth]{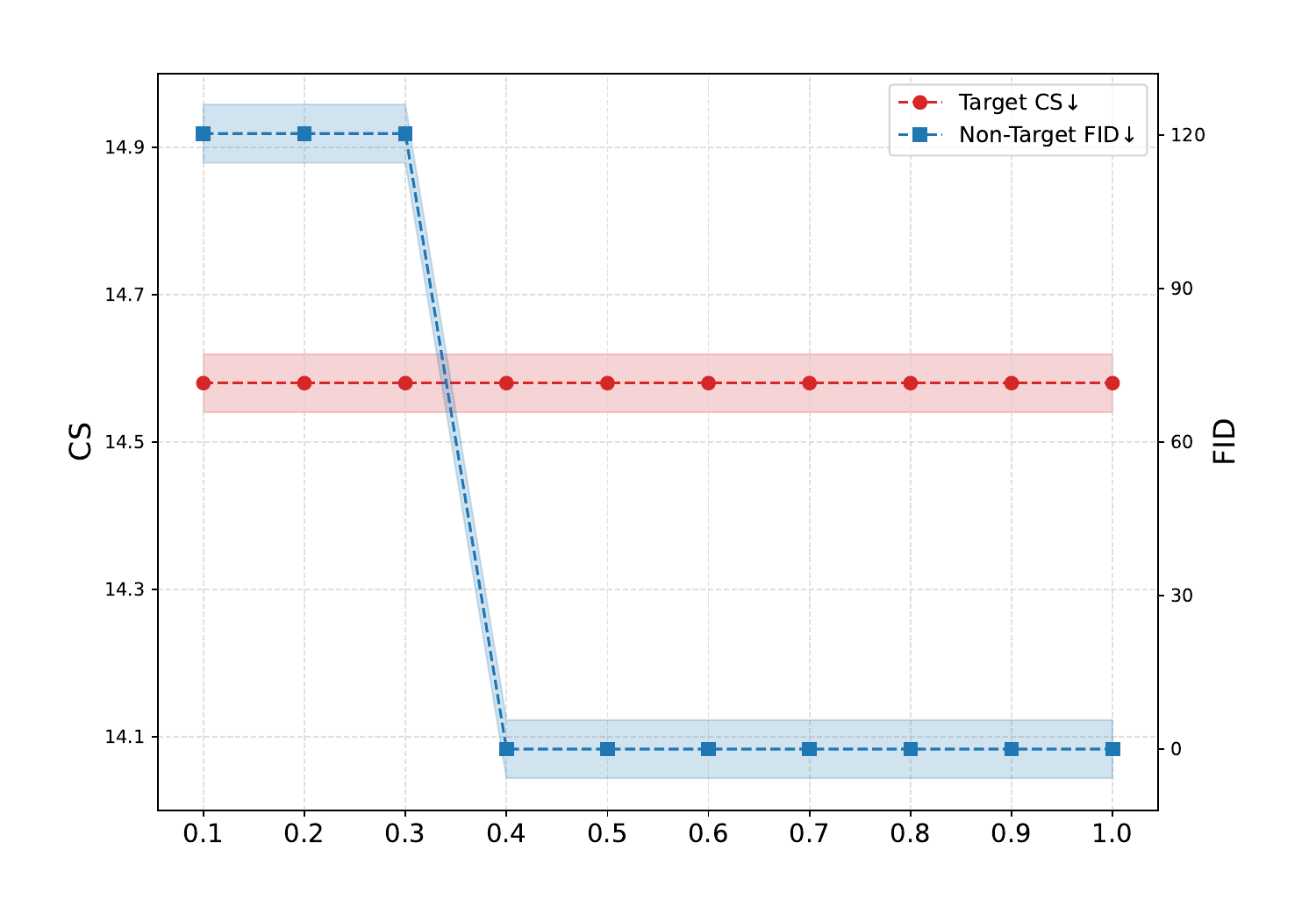}
    \caption{ $\gamma$}
    \label{fig:52}
  \end{subfigure}
  \hspace{0.01\linewidth}
  \begin{subfigure}{0.32\linewidth}
    \centering
    \includegraphics[width=\linewidth]{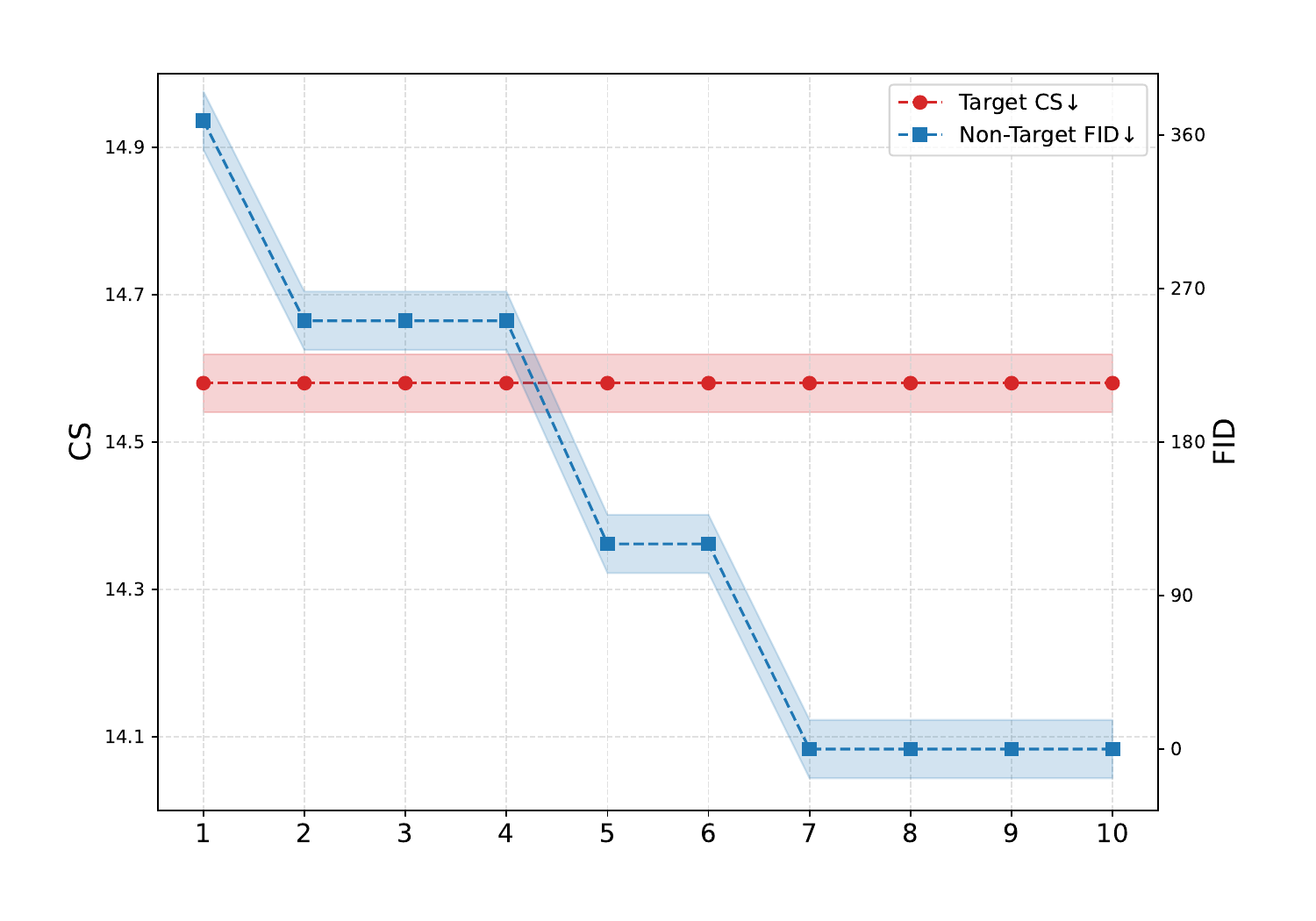}
    \caption{$\delta$}
    \label{fig:53}
  \end{subfigure}

  \caption{Effects of varying hyperparameters $K$, $\gamma$, and $\delta$ on erasing the concept Snoopy.}
  \label{fig:parameters}
\end{figure*}

\myparagraphsupp{This is supplementary to Section 5 ``\textbf{Experiments}''.} As shown in Figure~\ref{fig:parameters}, we perform ablation studies on three key GrOCE hyperparameters (cluster size $K$, decay factor $\gamma$, and projection threshold $\delta$) to assess their impact on single-concept erasure. The target CS remains consistently high across all settings, indicating stable erasure performance. For $K$, smaller values already capture the target semantic region, while larger values slightly degrade non-target preservation (FID). For $\gamma$, smaller values restrict propagation, whereas moderate to larger values achieve a better balance between erasure and preservation. For $\delta$, lower thresholds introduce interference to non-target concepts, while higher thresholds more precisely confine the erased region. Overall, these results demonstrate that GrOCE maintains robust and effective concept erasure across a wide range of hyperparameters while preserving unrelated content.

\end{document}


%% file: main.bbl
\begin{thebibliography}{45}
\providecommand{\natexlab}[1]{#1}
\providecommand{\url}[1]{\texttt{#1}}
\expandafter\ifx\csname urlstyle\endcsname\relax
  \providecommand{\doi}[1]{doi: #1}\else
  \providecommand{\doi}{doi: \begingroup \urlstyle{rm}\Url}\fi

\bibitem[Belrose et~al.(2023)Belrose, Schneider-Joseph, Ravfogel, Cotterell, Raff, and Biderman]{belrose2023leace}
Nora Belrose, David Schneider-Joseph, Shauli Ravfogel, Ryan Cotterell, Edward Raff, and Stella Biderman.
\newblock Leace: Perfect linear concept erasure in closed form.
\newblock \emph{Advances in Neural Information Processing Systems}, 36:\penalty0 66044--66063, 2023.

\bibitem[Biswas et~al.(2025)Biswas, Roy, and Roy]{biswas2025cure}
Shristi~Das Biswas, Arani Roy, and Kaushik Roy.
\newblock Cure: Concept unlearning via orthogonal representation editing in diffusion models.
\newblock In \emph{The Thirty-ninth Annual Conference on Neural Information Processing Systems}, 2025.

\bibitem[Bui et~al.(2024)Bui, Vuong, Doan, Le, Montague, Abraham, and Phung]{bui2024erasing}
Anh Bui, Long Vuong, Khanh Doan, Trung Le, Paul Montague, Tamas Abraham, and Dinh Phung.
\newblock Erasing undesirable concepts in diffusion models with adversarial preservation.
\newblock In \emph{Advances in Neural Information Processing systems}, pages 1--29, 2024.

\bibitem[Chen et~al.(2025)Chen, Guo, Wang, Zhang, Nie, and Liu]{chen2025trce}
Ruidong Chen, Honglin Guo, Lanjun Wang, Chenyu Zhang, Weizhi Nie, and An-An Liu.
\newblock Trce: Towards reliable malicious concept erasure in text-to-image diffusion models.
\newblock In \emph{Proceedings of the IEEE/CVF International Conference on Computer Vision}, pages 18927--18936, 2025.

\bibitem[Dhariwal and Nichol(2021)]{dhariwal2021diffusion}
Prafulla Dhariwal and Alexander Nichol.
\newblock Diffusion models beat gans on image synthesis.
\newblock \emph{Advances in Neural Information Processing Systems}, 34:\penalty0 8780--8794, 2021.

\bibitem[Fan et~al.(2024)Fan, Liu, Zhang, Wong, Wei, and Liu]{fan2023salun}
Chongyu Fan, Jiancheng Liu, Yihua Zhang, Eric Wong, Dennis Wei, and Sijia Liu.
\newblock Salun: Empowering machine unlearning via gradient-based weight saliency in both image classification and generation.
\newblock In \emph{International Conference on Learning Representations}, pages 1--31, 2024.

\bibitem[Gandikota et~al.(2023)Gandikota, Materzynska, Fiotto-Kaufman, and Bau]{gandikota2023erasing}
Rohit Gandikota, Joanna Materzynska, Jaden Fiotto-Kaufman, and David Bau.
\newblock Erasing concepts from diffusion models.
\newblock In \emph{Proceedings of the IEEE/CVF International Conference on Computer Vision}, pages 2426--2436, 2023.

\bibitem[Gandikota et~al.(2024)Gandikota, Orgad, Belinkov, Materzy{\'n}ska, and Bau]{gandikota2024unified}
Rohit Gandikota, Hadas Orgad, Yonatan Belinkov, Joanna Materzy{\'n}ska, and David Bau.
\newblock Unified concept editing in diffusion models.
\newblock In \emph{Proceedings of the IEEE/CVF Winter Conference on Applications of Computer Vision}, pages 5111--5120, 2024.

\bibitem[Gao et~al.(2025)Gao, Lu, Zhou, Chu, Zhang, Jia, Zhang, Fan, and Zhang]{gao2025eraseanything}
Daiheng Gao, Shilin Lu, Wenbo Zhou, Jiaming Chu, Jie Zhang, Mengxi Jia, Bang Zhang, Zhaoxin Fan, and Weiming Zhang.
\newblock Eraseanything: Enabling concept erasure in rectified flow transformers.
\newblock In \emph{Forty-second International Conference on Machine Learning}, 2025.

\bibitem[Gong et~al.(2024)Gong, Chen, Wei, Chen, and Jiang]{gong2024reliable}
Chao Gong, Kai Chen, Zhipeng Wei, Jingjing Chen, and Yu-Gang Jiang.
\newblock Reliable and efficient concept erasure of text-to-image diffusion models.
\newblock In \emph{European Conference on Computer Vision}, pages 73--88, 2024.

\bibitem[Han et~al.(2025)Han, Chen, Gong, Wei, Chen, and Jiang]{han2025dumo}
Feng Han, Kai Chen, Chao Gong, Zhipeng Wei, Jingjing Chen, and Yu-Gang Jiang.
\newblock Dumo: Dual encoder modulation network for precise concept erasure.
\newblock In \emph{Proceedings of the AAAI Conference on Artificial Intelligence}, pages 3320--3328, 2025.

\bibitem[Heng and Soh(2023)]{heng2023selective}
Alvin Heng and Harold Soh.
\newblock Selective amnesia: A continual learning approach to forgetting in deep generative models.
\newblock \emph{Advances in Neural Information Processing Systems}, 36:\penalty0 17170--17194, 2023.

\bibitem[Heusel et~al.(2017)Heusel, Ramsauer, Unterthiner, Nessler, and Hochreiter]{heusel2017gans}
Martin Heusel, Hubert Ramsauer, Thomas Unterthiner, Bernhard Nessler, and Sepp Hochreiter.
\newblock Gans trained by a two time-scale update rule converge to a local nash equilibrium.
\newblock \emph{Advances in Neural Information Processing Systems}, 30, 2017.

\bibitem[Ho et~al.(2020)Ho, Jain, and Abbeel]{ho2020denoising}
Jonathan Ho, Ajay Jain, and Pieter Abbeel.
\newblock Denoising diffusion probabilistic models.
\newblock \emph{Advances in Neural Information Processing Systems}, 33:\penalty0 6840--6851, 2020.

\bibitem[Huang et~al.(2024)Huang, Chang, Tsai, Lai, Yang, and Wang]{huang2024receler}
Chi-Pin Huang, Kai-Po Chang, Chung-Ting Tsai, Yung-Hsuan Lai, Fu-En Yang, and Yu-Chiang~Frank Wang.
\newblock Receler: Reliable concept erasing of text-to-image diffusion models via lightweight erasers.
\newblock In \emph{European Conference on Computer Vision}, pages 360--376, 2024.

\bibitem[Kumari et~al.(2023)Kumari, Zhang, Wang, Shechtman, Zhang, and Zhu]{kumari2023ablating}
Nupur Kumari, Bingliang Zhang, Sheng-Yu Wang, Eli Shechtman, Richard Zhang, and Jun-Yan Zhu.
\newblock Ablating concepts in text-to-image diffusion models.
\newblock In \emph{Proceedings of the IEEE/CVF International Conference on Computer Vision}, pages 22691--22702, 2023.

\bibitem[Labs(2024)]{flux2024}
Black~Forest Labs.
\newblock Flux.
\newblock \url{https://github.com/black-forest-labs/flux}, 2024.

\bibitem[Lee et~al.(2025{\natexlab{a}})Lee, Lim, and Chun]{lee2025localized}
Byung~Hyun Lee, Sungjin Lim, and Se~Young Chun.
\newblock Localized concept erasure for text-to-image diffusion models using training-free gated low-rank adaptation.
\newblock In \emph{Proceedings of the Computer Vision and Pattern Recognition Conference}, pages 18596--18606, 2025{\natexlab{a}}.

\bibitem[Lee et~al.(2025{\natexlab{b}})Lee, Lim, Lee, Kang, and Chun]{lee2025concept}
Byung~Hyun Lee, Sungjin Lim, Seunggyu Lee, Dong~Un Kang, and Se~Young Chun.
\newblock Concept pinpoint eraser for text-to-image diffusion models via residual attention gate.
\newblock \emph{arXiv:2506.22806}, 2025{\natexlab{b}}.

\bibitem[Leu et~al.(2024)Leu, Nakashima, and Garcia]{leu2024auditing}
Warren Leu, Yuta Nakashima, and Noa Garcia.
\newblock Auditing image-based nsfw classifiers for content filtering.
\newblock In \emph{Proceedings of the ACM Conference on Fairness, Accountability, and Transparency}, pages 1163--1173, 2024.

\bibitem[Li et~al.(2025{\natexlab{a}})Li, Zhang, Sun, and Yang]{li2025detect}
Feifei Li, Mi Zhang, Yiming Sun, and Min Yang.
\newblock Detect-and-guide: Self-regulation of diffusion models for safe text-to-image generation via guideline token optimization.
\newblock In \emph{Proceedings of the Computer Vision and Pattern Recognition Conference}, pages 13252--13262, 2025{\natexlab{a}}.

\bibitem[Li et~al.(2025{\natexlab{b}})Li, Xiao, Ji, Deng, Hui, Guo, and Ma]{li2025sculpting}
Gen Li, Yang Xiao, Jie Ji, Kaiyuan Deng, Bo Hui, Linke Guo, and Xiaolong Ma.
\newblock Sculpting memory: Multi-concept forgetting in diffusion models via dynamic mask and concept-aware optimization.
\newblock In \emph{Proceedings of the IEEE/CVF International Conference on Computer Vision}, pages 19659--19668, 2025{\natexlab{b}}.

\bibitem[Li et~al.(2026)Li, Wang, Hu, Jiang, Liang, Hao, Ma, and Feng]{li2025speed}
Ouxiang Li, Yuan Wang, Xinting Hu, Houcheng Jiang, Tao Liang, Yanbin Hao, Guojun Ma, and Fuli Feng.
\newblock {SPEED}: Scalable, precise, and efficient concept erasure for diffusion models.
\newblock In \emph{International Conference on Learning Representations}, pages 1--27, 2026.

\bibitem[Li et~al.(2024)Li, van~de Weijer, Khan, Hou, Wang, et~al.]{li2024get}
Senmao Li, Joost van~de Weijer, Fahad Khan, Qibin Hou, Yaxing Wang, et~al.
\newblock Get what you want, not what you don't: Image content suppression for text-to-image diffusion models.
\newblock In \emph{International Conference on Learning Representations}, pages 1--27, 2024.

\bibitem[Lu et~al.(2022)Lu, Zhou, Bao, Chen, Li, and Zhu]{lu2022dpm}
Cheng Lu, Yuhao Zhou, Fan Bao, Jianfei Chen, Chongxuan Li, and Jun Zhu.
\newblock Dpm-solver: A fast ode solver for diffusion probabilistic model sampling in around 10 steps.
\newblock \emph{Advances in Neural Information Processing Systems}, 35:\penalty0 5775--5787, 2022.

\bibitem[Lu et~al.(2024)Lu, Wang, Li, Liu, and Kong]{lu2024mace}
Shilin Lu, Zilan Wang, Leyang Li, Yanzhu Liu, and Adams Wai-Kin Kong.
\newblock Mace: Mass concept erasure in diffusion models.
\newblock In \emph{Proceedings of the IEEE/CVF Conference on Computer Vision and Pattern Recognition}, pages 6430--6440, 2024.

\bibitem[Lyu et~al.(2024)Lyu, Yang, Hong, Chen, Jin, He, Xue, Han, and Ding]{lyu2024one}
Mengyao Lyu, Yuhong Yang, Haiwen Hong, Hui Chen, Xuan Jin, Yuan He, Hui Xue, Jungong Han, and Guiguang Ding.
\newblock One-dimensional adapter to rule them all: Concepts diffusion models and erasing applications.
\newblock In \emph{Proceedings of the IEEE/CVF Conference on Computer Vision and Pattern Recognition}, pages 7559--7568, 2024.

\bibitem[Nichol and Dhariwal(2021)]{nichol2021improved}
Alexander~Quinn Nichol and Prafulla Dhariwal.
\newblock Improved denoising diffusion probabilistic models.
\newblock In \emph{International Conference on Machine Learning}, pages 8162--8171, 2021.

\bibitem[Orgad et~al.(2023)Orgad, Kawar, and Belinkov]{orgad2023editing}
Hadas Orgad, Bahjat Kawar, and Yonatan Belinkov.
\newblock Editing implicit assumptions in text-to-image diffusion models.
\newblock In \emph{Proceedings of the IEEE/CVF International Conference on Computer Vision}, pages 7053--7061, 2023.

\bibitem[Podell et~al.(2023)Podell, English, Lacey, Blattmann, Dockhorn, M{\"u}ller, Penna, and Rombach]{podell2023sdxl}
Dustin Podell, Zion English, Kyle Lacey, Andreas Blattmann, Tim Dockhorn, Jonas M{\"u}ller, Joe Penna, and Robin Rombach.
\newblock Sdxl: Improving latent diffusion models for high-resolution image synthesis.
\newblock In \emph{International Conference on Learning Representations}, pages 1--18, 2023.

\bibitem[Radford et~al.(2021)Radford, Kim, Hallacy, Ramesh, Goh, Agarwal, Sastry, Askell, Mishkin, Clark, et~al.]{radford2021learning}
Alec Radford, Jong~Wook Kim, Chris Hallacy, Aditya Ramesh, Gabriel Goh, Sandhini Agarwal, Girish Sastry, Amanda Askell, Pamela Mishkin, Jack Clark, et~al.
\newblock Learning transferable visual models from natural language supervision.
\newblock In \emph{International Conference on Machine Learning}, pages 8748--8763, 2021.

\bibitem[Rombach et~al.(2022)Rombach, Blattmann, Lorenz, Esser, and Ommer]{rombach2022high}
Robin Rombach, Andreas Blattmann, Dominik Lorenz, Patrick Esser, and Bj{\"o}rn Ommer.
\newblock High-resolution image synthesis with latent diffusion models.
\newblock In \emph{Proceedings of the IEEE/CVF Conference on Computer Vision and Pattern Recognition}, pages 10684--10695, 2022.

\bibitem[Schramowski et~al.(2023)Schramowski, Brack, Deiseroth, and Kersting]{schramowski2023safe}
Patrick Schramowski, Manuel Brack, Bj{\"o}rn Deiseroth, and Kristian Kersting.
\newblock Safe latent diffusion: Mitigating inappropriate degeneration in diffusion models.
\newblock In \emph{Proceedings of the IEEE/CVF Conference on Computer Vision and Pattern Recognition}, pages 22522--22531, 2023.

\bibitem[Somepalli et~al.(2023)Somepalli, Singla, Goldblum, Geiping, and Goldstein]{somepalli2023diffusion}
Gowthami Somepalli, Vasu Singla, Micah Goldblum, Jonas Geiping, and Tom Goldstein.
\newblock Diffusion art or digital forgery? investigating data replication in diffusion models.
\newblock In \emph{Proceedings of the IEEE/CVF Conference on Computer Vision and Pattern Recognition}, pages 6048--6058, 2023.

\bibitem[Stanczuk et~al.(2024)Stanczuk, Batzolis, Deveney, and Sch{\"o}nlieb]{stanczuk2024diffusion}
Jan~Pawel Stanczuk, Georgios Batzolis, Teo Deveney, and Carola-Bibiane Sch{\"o}nlieb.
\newblock Diffusion models encode the intrinsic dimension of data manifolds.
\newblock In \emph{Forty-first International Conference on Machine Learning}, 2024.

\bibitem[Thakral et~al.(2025)Thakral, Glaser, Hassner, Vatsa, and Singh]{thakral2025fine}
Kartik Thakral, Tamar Glaser, Tal Hassner, Mayank Vatsa, and Richa Singh.
\newblock Fine-grained erasure in text-to-image diffusion-based foundation models.
\newblock In \emph{Proceedings of the IEEE/CVF Conference on Computer Vision and Pattern Recognition}, pages 9121--9130, 2025.

\bibitem[Wang et~al.(2025)Wang, Li, Mu, Hao, Liu, Wang, and He]{wang2025precise}
Yuan Wang, Ouxiang Li, Tingting Mu, Yanbin Hao, Kuien Liu, Xiang Wang, and Xiangnan He.
\newblock Precise, fast, and low-cost concept erasure in value space: Orthogonal complement matters.
\newblock In \emph{Proceedings of the Computer Vision and Pattern Recognition Conference}, pages 28759--28768, 2025.

\bibitem[Yang et~al.(2022)Yang, Li, Dai, and Gao]{yang2022focal}
Jianwei Yang, Chunyuan Li, Xiyang Dai, and Jianfeng Gao.
\newblock Focal modulation networks.
\newblock \emph{Advances in Neural Information Processing Systems}, 35:\penalty0 4203--4217, 2022.

\bibitem[Yang et~al.(2024)Yang, Hui, Yuan, Gong, and Cao]{yang2024sneakyprompt}
Yuchen Yang, Bo Hui, Haolin Yuan, Neil Gong, and Yinzhi Cao.
\newblock Sneakyprompt: Jailbreaking text-to-image generative models.
\newblock In \emph{IEEE Symposium on Security and Privacy}, pages 897--912, 2024.

\bibitem[Zeng et~al.(2025)Zeng, Cao, Cao, Chang, Chen, and Lin]{zeng2024advi2i}
Yaopei Zeng, Yuanpu Cao, Bochuan Cao, Yurui Chang, Jinghui Chen, and Lu Lin.
\newblock Advi2i: Adversarial image attack on image-to-image diffusion models.
\newblock In \emph{International Conference on Machine Learning}, pages 1--14, 2025.

\bibitem[Zhang et~al.(2024{\natexlab{a}})Zhang, Wang, Xu, Wang, and Shi]{zhang2024forget}
Gong Zhang, Kai Wang, Xingqian Xu, Zhangyang Wang, and Humphrey Shi.
\newblock Forget-me-not: Learning to forget in text-to-image diffusion models.
\newblock In \emph{Proceedings of the IEEE/CVF Conference on Computer Vision and Pattern Recognition}, pages 1755--1764, 2024{\natexlab{a}}.

\bibitem[Zhang et~al.(2023)Zhang, Rao, and Agrawala]{zhang2023adding}
Lvmin Zhang, Anyi Rao, and Maneesh Agrawala.
\newblock Adding conditional control to text-to-image diffusion models.
\newblock In \emph{Proceedings of the IEEE/CVF International Conference on Computer Vision}, pages 3836--3847, 2023.

\bibitem[Zhang et~al.(2024{\natexlab{b}})Zhang, Chen, Jia, Zhang, Fan, Liu, Hong, Ding, and Liu]{zhang2024defensive}
Yimeng Zhang, Xin Chen, Jinghan Jia, Yihua Zhang, Chongyu Fan, Jiancheng Liu, Mingyi Hong, Ke Ding, and Sijia Liu.
\newblock Defensive unlearning with adversarial training for robust concept erasure in diffusion models.
\newblock \emph{Advances in Neural Information Processing Systems}, 37:\penalty0 36748--36776, 2024{\natexlab{b}}.

\bibitem[Zhang et~al.(2024{\natexlab{c}})Zhang, Jia, Chen, Chen, Zhang, Liu, Ding, and Liu]{zhang2024generate}
Yimeng Zhang, Jinghan Jia, Xin Chen, Aochuan Chen, Yihua Zhang, Jiancheng Liu, Ke Ding, and Sijia Liu.
\newblock To generate or not? safety-driven unlearned diffusion models are still easy to generate unsafe images... for now.
\newblock In \emph{European Conference on Computer Vision}, pages 385--403, 2024{\natexlab{c}}.

\bibitem[Zhu et~al.(2025)Zhu, Cui, Yan, and Han]{zhu2025reinforcing}
Hegui Zhu, Wenqi Cui, Yue Yan, and Ning Han.
\newblock Reinforcing adversarial transferability via negative class guided example generation.
\newblock \emph{IEEE Transactions on Information Forensics and Security}, 21:\penalty0 532--546, 2025.

\end{thebibliography}
